\documentclass[10pt,journal,compsoc]{IEEEtran}

\usepackage[pdftex]{graphicx}
\usepackage{amsopn}
\usepackage{booktabs} 
\usepackage{url}
\usepackage{color}
\usepackage{multirow}
\usepackage{color}
\usepackage[switch]{lineno}
\usepackage{graphicx}
\usepackage{subfigure}
\usepackage{url}
\usepackage{epstopdf}
\usepackage{hyperref}
\usepackage[table,xcdraw]{xcolor}
\usepackage{amsmath}
\usepackage{textcomp}
\usepackage{verbatim}
\usepackage{bm}

\usepackage{amssymb}
\usepackage{amsthm}
\usepackage{amsfonts}
\usepackage{enumitem}
\usepackage{ulem}

\theoremstyle{definition}
\newtheorem{myDef}{Definition}

\ifCLASSOPTIONcompsoc
  \usepackage[nocompress]{cite}
\else
  \usepackage{cite}
\fi

\ifCLASSINFOpdf
\else
\fi

\begin{document}

\title{Online Metro Origin-Destination Prediction via Heterogeneous Information Aggregation}

\author{Lingbo Liu,
        Yuying Zhu,
        Guanbin Li,
        Ziyi Wu,
        Lei Bai,
        and Liang Lin, {\textit{Senior Member, IEEE}}

\IEEEcompsocitemizethanks{
\IEEEcompsocthanksitem This work was supported in part by the National Key Research and Development Project of China under Grant No.2021ZD0111600, in part by the National Natural Science Foundation of China under Grant No.U1811463, No.61976250 and No.61836012, in part by the Guangdong Basic and Applied Basic Research Foundation under Grant No.2020B1515020048, in part by the National High-Level Talents Special Support Plan (Ten Thousand Talents Program), and in part by the Guangdong Science and Technology Project under Grant No.202102020633 and No.2017A030312006. (\textit{Corresponding Author: Liang Lin.})
\IEEEcompsocthanksitem L. Liu, Y. Zhu, G. Li, Z. Wu and L. Lin are with the School of Computer Science and Engineering, Sun Yat-Sen University, China, 510000 (e-mail: liulingb@mail2.sysu.edu.cn; zhuyy76@mail2.sysu.edu.cn; liguanbin@mail.sysu.edu.cn; wuzy39@mail2.sysu.edu.cn; linliang@ieee.org).
\IEEEcompsocthanksitem L. Lin is also with Key Laboratory of Machine Intelligence and Advanced Computing, Ministry of Education, and with Engineering Research Center for Advanced Computing Engineering Software of Ministry of Education, China. L. Lin is also with GuangDong Province Key Laboratory of Information Security Technology.
\IEEEcompsocthanksitem L. Bai is with the School of Electrical and Information Engineering, the University of Sydney, Australia 2000 (e-mail: lei.bai@sydney.edu.au).
}


}

\markboth{IEEE Transactions on Pattern Analysis and Machine Intelligence}%
{Liu \MakeLowercase{\textit{et al.}}: Online Metro Origin-Destination Ridership Prediction}

\IEEEtitleabstractindextext{%
\begin{abstract}
Metro origin-destination prediction is a crucial yet challenging time-series analysis task in intelligent transportation systems, which aims to accurately forecast two specific types of cross-station ridership, i.e., Origin-Destination (OD) one and Destination-Origin (DO) one. However, complete OD matrices of previous time intervals can not be obtained immediately in online metro systems, and conventional methods only used limited information to forecast the future OD and DO ridership separately.
In this work, we proposed a novel neural network module termed Heterogeneous Information Aggregation Machine (HIAM), which fully exploits heterogeneous information of historical data (e.g., incomplete OD matrices, unfinished order vectors, and DO matrices) to jointly learn the evolutionary patterns of OD and DO ridership. Specifically, an OD modeling branch estimates the potential destinations of unfinished orders explicitly to complement the information of incomplete OD matrices, while a DO modeling branch takes DO matrices as input to capture the spatial-temporal distribution of DO ridership. Moreover, a Dual Information Transformer is introduced to propagate the mutual information among OD features and DO features for modeling the OD-DO causality and correlation. Based on the proposed HIAM, we develop a unified Seq2Seq network to forecast the future OD and DO ridership simultaneously. Extensive experiments conducted on two large-scale benchmarks demonstrate the effectiveness of our method for online metro origin-destination prediction. Our code is resealed at {\color{blue}\url{https://github.com/HCPLab-SYSU/HIAM}}.
\end{abstract}

}

\maketitle

\IEEEdisplaynontitleabstractindextext
\IEEEpeerreviewmaketitle

\IEEEraisesectionheading{\section{Introduction}\label{sec:introduction}}
\IEEEPARstart{T}{ime} series prediction \cite{chen2021bayesian,spadondesouza2021pay} is one of the most active research topics in artificial intelligence. In this work, we pay attention to its practical application in transportation management, e.g., improving the operation efficiency of urban metro systems, since metro has become a popular travel mode for residents. It was reported that over 10 million metro travel transactions are made per day in some metropolises (e.g., Beijing and Shanghai) \cite{BJSubway,SHSubway}. Such a huge ridership poses great challenges for metro operation. In this case, accurately forecasting the future ridership is crucial for metro scheduling and route planning.

\begin{figure}[t]
    \centering
    \includegraphics[width=0.95\columnwidth]{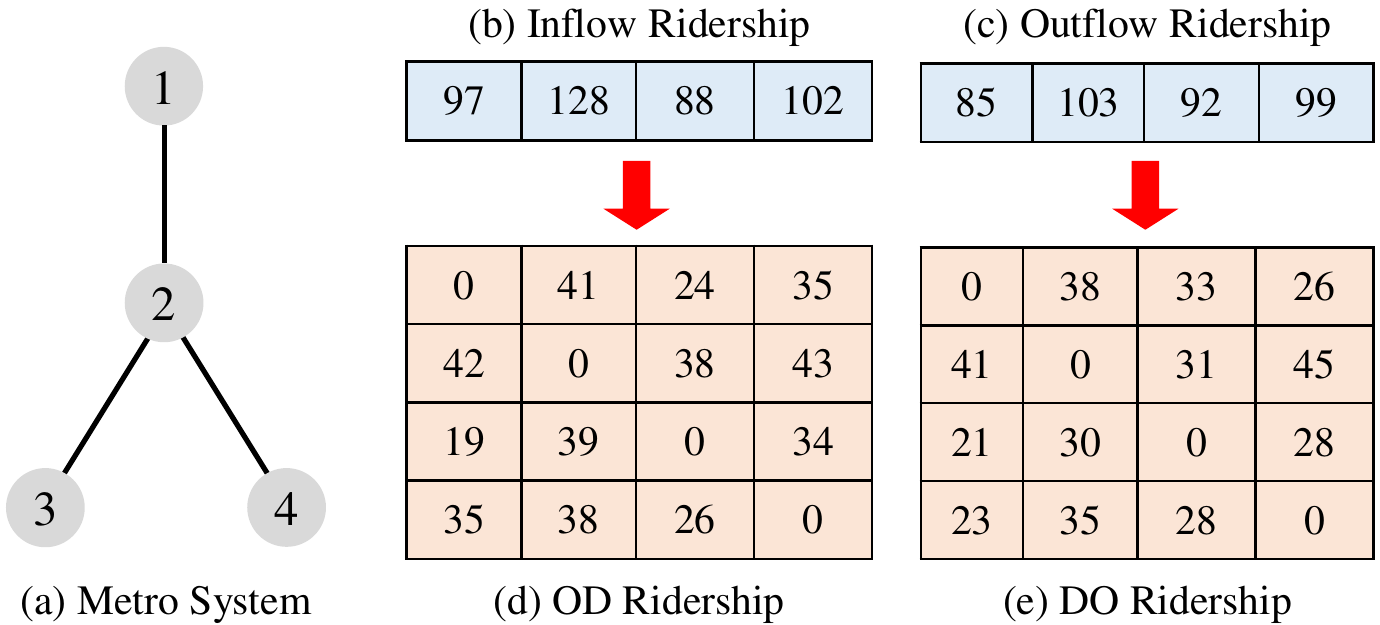}
    \vspace{-3mm}
    \caption{Illustration of the difference between station-level ridership and origin-destination ridership. (a) is a metro system with four stations. (b) and (c) are the inflow/outflow of each station, respectively. (d) is an Origin-Destination (OD) matrix that represents the destination distribution of incoming passengers. (e) is a Destination-Origin (DO) matrix that represents the origin distribution of outgoing passengers.}
    \label{fig:OD_ridership}
    \vspace{0mm}
\end{figure}

Due to its significant applications, metro ridership prediction has recently attracted extensive attention in both academic and industrial communities \cite{ma2018parallel,fang2019gstnet,hao2019sequence,liu2020physical,li2020tensor}. However, most conventional works were merely proposed for station-level prediction, i.e., forecasting the inflow and outflow of each metro station, as shown in Fig. \ref{fig:OD_ridership}-(b,c). Such information about inflow/outflow ridership is too coarse to reflect the mobility of passengers.
To explore more valuable information for metro optimization, we focus on a more challenging task, i.e., metro origin-destination prediction, whose goal is to forecast the ridership between any two stations over the next several time intervals. Specifically, two special types of cross-station ridership are taken into consideration in our work:
\begin{itemize}
\item {\bf{Origin-Destination (OD) Ridership:}} For each station, we aim to forecast the number of passengers entering at time interval $t$ and the stations they will go to. For example, the OD ridership at time interval $t$ can be represented as a matrix\footnote{OD matrix and DO matrix may be sparse since the ridership between some stations is usually small or even zero. In this work, we would compress these matrices by merging the small cross-station ridership. More details can be referred to Section \ref{sec:preliminary}. \label{matrix_real_size}} ${OD}_t \in \mathbb{R}^{N \times N}$, where $N$ is the total number of stations. More specifically, $OD_t(i,j)$ denotes the number of passengers that entered station $i$ at time interval $t$ and would head for station $j$, as shown in Fig. \ref{fig:OD_ridership}-(d).

\item {\bf{Destination-Origin (DO) Ridership:}} We also aim to predict the future outgoing ridership of each station and where these passengers come from. Similarly, the DO ridership at time interval $t$ can be represented as a matrix\textsuperscript{\ref {matrix_real_size}} $DO_t \in \mathbb{R}^{N \times N}$,  as shown in Fig. \ref{fig:OD_ridership}-(e). Specifically, $DO_t(i,j)$ is the number of passengers that entered station $j$ at earlier moments and exit from station $i$ at time interval $t$.
\end{itemize}

\begin{figure}[t]
    \centering
    \includegraphics[width=0.875\columnwidth]{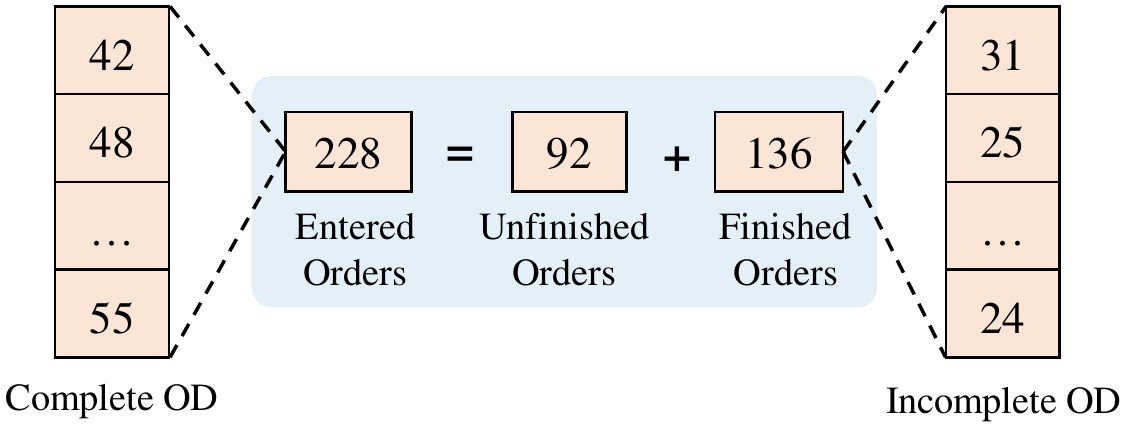}
    \vspace{-3mm}
    \caption{Illustration of the incomplete OD matrix in online metro systems. Suppose there were 228 passengers that entered station $\bm{s}$ in the past 15 minutes and 136 people have arrived at their destinations up to now. Unfortunately, the destinations of the remaining people are unknowable. In this case, we can only construct an incomplete OD matrix from finished trip transactions.}
    \label{fig:Incomplete_OD}
    \vspace{0mm}
\end{figure}

Intuitively, we should utilize the historical OD/DO ridership to forecast the future OD/DO ridership. Unfortunately, in online metro systems, the complete historical OD matrices can not be constructed in real time. One example is illustrated in Fig. \ref{fig:Incomplete_OD}. Suppose 228 passengers entered station $i$ in the past 15 minutes and 136 people have arrived at their destinations up to now. However, the destinations of the remaining ongoing passengers are unknowable, until they arrive at their exited stations. Under this circumstance, we can only construct an incomplete OD matrix based on the finished trip transactions. When most passengers are still on their way to destinations, such an incomplete matrix is very sparse and uninformative, which greatly increases the difficulties of metro origin-destination distribution modeling.

In literature, there are very few methods \cite{gong2018network,zhang2020short,gong2020online,noursalehi2021dynamic,cheng2021real} proposed for online metro origin-destination ridership prediction. Conventional works either take the incomplete historical OD matrices as input or directly utilize DO matrices to forecast the future OD matrices. Despite certain progress, these methods suffer from the following limitations. {\bf{First}}, these works don't explore the information about unfinished/ongoing trips. Intuitively, human mobilities are usually periodic \cite{zhang2017deep,yao2019revisiting} and we can estimate the potential destinations of those ongoing trips to some extent. Therefore, more information is available for metro OD prediction.
{\bf{Second}}, it is sub-optimal to directly use the historical DO information to toughly forecast the future OD matrices \cite{noursalehi2021dynamic}, since the former isn't the essential factor affecting the evolution of the latter. {\bf{Third}}, all above-mentioned methods are unaware of the mutual information between OD and DO ridership, i.e., forecasting the future OD and DO matrices separately. In essence, the previous OD ridership would greatly influence the future DO ridership, which is called OD-to-DO causality in our work. Moreover, there also exists a relationship between the previous DO ridership and the future OD ridership. For example, the DO and OD ridership of tide stations in residential and office areas are usually negatively correlated \cite{gong2008data}. Such a relationship is called DO-to-OD correlation. In summary, previous methods only exploit limited information and cannot effectively model the metro ridership distribution.

To tackle the aforementioned problems, we propose a unified neural network module termed Heterogeneous Information Aggregation Machine (HIAM), which fully aggregates heterogeneous information of historical ridership to learn the evolutionary trend of future origin-destination ridership. In particular, our HIAM consists of {\bf i)} two parallel branches respectively for OD and DO modeling, and {\bf ii)} a Dual Information Transformer for OD-DO interaction modeling.
Unlike previous works \cite{gong2018network,gong2020online} that neglected unfinished transactions, we exploit the information of these transactions explicitly to complement the incomplete OD matrices. Specifically, our OD branch explores the long short-term historical destination distribution to estimate two potential destination matrices of ongoing passengers, which are incorporated with the incomplete OD matrix and fed into graph convolutional gated recurrent units (GCGRU) to generate a compact OD hidden state. Meanwhile, the DO branch feeds the corresponding DO matrix into a GCGRU for DO hidden state generation. To model the internal interaction among OD and DO ridership, our Dual Information Transformer enhances the OD state and DO state mutually by propagating their complementary information in a dual manner. These refined hidden states are respectively fed into the following GCGRU for high-order spatial-temporal representation learning.
Based on the tailor-designed HIAM, we develop a unified online metro origin-destination prediction framework with a Seq2Seq architecture \cite{sutskever2014sequence}, which jointly forecasts the OD ridership and DO ridership of the next several time intervals. Finally, we conduct extensive experiments on two large-scale benchmarks (i.e., Shanghai Metro and Hangzhou Metro), and evaluation results show that our approach outperforms existing state-of-the-art methods for both OD prediction and DO prediction.

In summary, the contributions of this work are four-fold:
\begin{itemize}
\item We propose a novel Heterogeneous Information Aggregation Machine to facilitate the online metro origin-destination prediction. To the best of our knowledge, our HIAM is the first deep learning approach that fully aggregates heterogeneous information of incomplete OD matrices, unfinished order vectors, and DO matrices to forecast the future cross-station ridership.

\item To fully exploit the information of unfinished transactions, our HIAM explores the long short-term historical distribution to estimate the potential destinations of ongoing passengers, which are further utilized to complement the information of incomplete OD matrices.

\item A Dual Information Transformer is introduced to propagate the mutual information among OD features and DO features, thus better modeling the internal interaction between OD and DO ridership. To the best of our best knowledge, our work is the first attempt to employ the heterogeneous transformer to address time series forecasting.

\item Extensive experiments conducted on two large-scale datasets show the effectiveness of the proposed method for both OD ridership prediction and DO ridership prediction of online metro systems.
\end{itemize}

The rest of this paper is organized as follows. First, we review some related works of traffic state prediction and origin-destination prediction in Section \ref{sec:review}. We then provide some preliminaries in Section \ref{sec:preliminary} and introduce the proposed approach for online metro prediction in Section \ref{sec:method}. Extensive comparison and ablation analysis are conducted in Section~\ref{sec:experiment}. Finally, we conclude this paper in Section~\ref{sec:conclusion}.

\section{Related Works}\label{sec:review}
\subsection{Traffic Time Series Prediction}
Accurate prediction of future traffic states is a crucial task of time series analysis and it has widespread applications in intelligent transportation systems. In literature, a large number of methods \cite{lin2017road,zhang2019flow,pan2020spatio,zheng2020gman,tedjopurnomo2020survey,yin2020comprehensive,guo2021learning} have been proposed to address this task. Early works usually applied time series models for prediction, but could not well model the traffic patterns of complex and unconstrained scenarios  \cite{lippi2013short,guo2014adaptive,dell2015time}.
Recently, deep neural networks have become the mainstream approach in this field. For instance, Wang et al. \cite{wang2017deepsd} developed an end-to-end convolutional neural network to automatically discover the supply-demand patterns from car-hailing service data. Zhang \textit{et al.} \cite{zhang2017deep} utilized three residual networks \cite{he2016deep} to learn the closeness, period and trend properties for citywide traffic flow prediction. Yao \textit{et al.} \cite{yao2018deep} proposed a Deep Multi-View Spatial-Temporal Network for taxi demand prediction, which learned spatial relations with a deep CNN and modeled temporal correlations with a Long Short-Term Memory (LSTM) cell \cite{xingjian2015convolutional}. Liu \textit{et al.} \cite{liu2018attentive,liu2020dynamic} incorporated a hierarchical convolutional LSTM network with an attention mechanism to learn the spatial-temporal representations dynamically.

Subsequently, graph convolutional networks (GCN \cite{bruna2014spectral,duvenaud2015convolutional,chen2020knowledge,chen2021cross}) were widely adopted to model various complex systems with non-Euclidean structures. For instance, Li \textit{et al.} \cite{li2018diffusion} modeled the traffic flow as a diffusion process on a directed graph and captured the spatial dependency with bidirectional random walks. Guo \textit{et al.} \cite{guo2019attention} introduced attention mechanisms into spatial-temporal graph networks for dynamical traffic prediction. Bai \textit{et al.} \cite{bai2019stg2seq} utilized a hierarchical graph convolutional structure to capture both the spatial and temporal correlations for multi-step passenger demand prediction. Sun \textit{et al.} \cite{sun2020predicting} developed a multi-view graph convolutional network for crowd flow prediction, where different views captured various interactions and spatial correlations between different irregular regions. Cao \textit{et al.} \cite{cao2020spectral} proposed a Spectral Temporal Graph Neural Network, which incorporated a Graph Fourier Transform and a Discrete Fourier Transform to jointly capture inter-series correlations and temporal dependencies in the spectral domain.
Recently, GCN has also been employed for metro ridership prediction. Han \textit{et al.} \cite{han2019predicting} transformed the city metro network into a graph and made predictions using graph convolutional neural networks. Liu \textit{et al.} \cite{liu2020physical} modeled a metro system as graphs with physical/virtual topologies and proposed a unified Physical-Virtual Collaboration Graph Network, which fully learned the complex ridership patterns from those tailor-designed graphs.
Nevertheless, most previous methods merely forecast the traffic states of every region or station, which can only provide limited information for urban traffic management. Thus we focus on the more meaningful origin-destination prediction in this work.

\subsection{Traffic Origin-Destination Prediction}
Traffic origin-destination prediction is a challenging task that aims to forecast the traffic flow or demand between any two positions. Recently, this task has attracted increasing attention in both academic and industrial communities.
For instance, Liu \textit{et al.} \cite{liu2019contextualized} incorporated local spatial context, temporal evolution context, and global correlation context to forecast the future taxi OD demand. Shi \textit{et al.} \cite{shi2020predicting} extracted temporal features for each OD pair with LSTM units and learned the spatial dependency of origins and destinations respectively. Wang \textit{et al.} \cite{wang2019origin} applied graph convolutions among geographical and semantic neighbors to model the taxi OD transferring patterns. Ke \textit{et al.} \cite{ke2021predicting} characterized the OD pair-wise relationships with multiple OD graphs and developed a spatial-temporal encoder-decoder residual framework to model both the spatial dependencies across different OD pairs and the temporal dependencies of the OD pairs themselves.

All methods mentioned above were proposed for ride-hailing applications, where both the origin and destination of a passenger are known once a taxi request is generated. However, in online metro systems, the destination of a passenger is unknowable until he/she reaches the destination station, thus the complete OD matrices cannot be obtained immediately. Thus online metro origin-destination prediction essentially belongs to the category of incomplete time series analysis \cite{ma2019end,ma2020adversarial}. To address this problem, Gong \textit{et al.} \cite{gong2020online} used some indication matrices to mask and neglect those unfinished metro orders and applied a non-negative matrix factorization strategy to learn the latent properties of entered and exited stations from incomplete OD matrices. Both Zhang \textit{et al.} \cite{zhang2020short} and Cheng \textit{et al.} \cite{cheng2021real} used the boarding (entering) demand to replace the unavailable OD matrices. Specifically, Zhang \textit{et al.} \cite{zhang2020short} developed a channel-wise attentive split–convolutional neural network to assign different values for OD features, while Cheng \textit{et al.} \cite{cheng2021real} developed a high-order weighted dynamic mode decomposition to learn time-evolving features of a metro system. Recently, Noursalehi \textit{et al.} \cite{noursalehi2021dynamic} toughly used the historical DO matrices to forecast the future OD matrices with a multi-resolution spatial-temporal neural network model.
However, all previous works have never explicitly exploited the information of unfinished transactions. Moreover, they either only predicted OD matrices or forecasted OD/DO matrices separately, completely neglecting their intrinsic correlation. In contrast, our method fully aggregates various information to jointly forecast the future OD ridership and DO ridership.

\subsection{Transformer Architecture}

Transformer \cite{vaswani2017attention} is an advanced neural network block that aggregates information from the entire input sequence with an attention mechanism \cite{bahdanau2014neural}. Specifically, it consists of a multi-head self-attention layer, a point-wise feed-forward layer, and a normalization layer. The global computation and perfect memory mechanism make it suitable for long sequence modeling. Recently, transformer has been widely applied to various tasks of artificial intelligence, inducing natural language processing \cite{devlin2018bert,zhang2020sg}, computer vision \cite{parmar2018image,dosovitskiy2020image,liu2021road}, and time series analysis \cite{xu2020spatial,wu2020deep}. For instance, Zhou \textit{et al.} \cite{zhou2021informer} proposed an efficient transformer-based model, which incorporated a ProbSparse Self-attention mechanism and distilling operation to capture time-series long-range dependencies between outputs and inputs efficiently. Inspired by the success of these works, we apply a transformer to learn the global information interaction between all metro stations. However, unlike most previous works that only transferred homogeneous information, our method develops a Dual Information Transformer, which propagates OD information and DO information mutually by generating heterogeneous queries, keys, and values. To the best of our knowledge, our work is the first attempt to employ the heterogeneous transformer to address time series prediction.

\section{Preliminaries}\label{sec:preliminary}
In this section, we briefly introduce some notations and the definition for online metro origin-destination prediction. For brevity, the frequently used notations in this paper are summarized in Table \ref{tab:Notations}.

{\bf{1) Transaction Data:}} In an online metro system, a large number of trip transactions are made over time. For each finished transaction, we can know both its entry station and exit station, as well as their corresponding time-stamps. However, for each ongoing transaction, only the entry station and entry time-stamp are knowable.

{\bf{2) Matrix Compression Preprocessing:}} It's worth noting that the ridership between some stations is usually small or even zero. Inspired by the previous work \cite{liu2020physical}, we would take into consideration the origin-destination pairs that have high ridership.
Specifically, for each origin station, we measure the destination distribution of all its entered passengers and then select the top $K-1$ stations with the highest destination probability. Thus, we can generate an OD mapping matrix $M_{od} \in \mathbb{R}^{N \times K}$, where $N$ is the number of metro stations. More specifically, for the origin station $i$, $M_{od}(i,j)$ $(j=1,...,K-1)$ is the index of its $j$-th destination station with high ridership. Moreover, $M_{od}(i,K)$ is set to -1, which is utilized to indicate the indexes of the remaining destination stations. In the same way, we can also generate a DO mapping matrix $M_{do} \in \mathbb{R}^{N \times K}$, where $M_{do}(i,j)$ is the index of the $j$-th most related origin station for the destination station $i$. With a value of -1, $M_{do}(i,K)$ denotes the indexes of remaining origin stations. In this work, these mapping matrices are utilized to guide the compression of metro OD and DO matrices.

\begin{table}[t]
 \caption{Some notations for online metro origin-destination ridership prediction.}
  \vspace{-2mm}
  \newcommand{\tabincell}[2]{\begin{tabular}{@{}#1@{}}#2\end{tabular}}
  \centering
  \resizebox{9cm}{!} {
    \begin{tabular}{l|l}
    \hline
    {\textbf{Notations}} & {\textbf{Description}} \\
    \hline
    \hline
    $t$ & the current time interval\\
    \hline
    $n$ &  the length of input historical sequence \\
     \hline
    $m$ & the length of output future sequence \\
    \hline
    $N$ & the number of metro stations\\
    \hline
    $K$ &  \tabincell{l}{the number of considered OD/DO pairs\\for each station}\\
    \hline
    $M_{od} \in \mathbb{R}^{N{\times}K}$ & the index of considered OD pairs\\
    \hline
    $M_{do} \in \mathbb{R}^{N{\times}K}$ & the index of considered DO pairs\\
    \hline
    $U_{t-n+i} \in \mathbb{R}^{N}$ & the unfinished order vector {\color{white}.}($i=1,...,n$)\\
    \hline
    $IOD_{t-n+i} \in \mathbb{R}^{N{\times}K}$ & the incomplete OD matrix {\color{white}...}($i=1,...,n$) \\
    \hline
    $OD_{t+i} \in \mathbb{R}^{N{\times}K}$ & the complete OD matrix {\color{white}.......}($i=1,...,m$)\\
    \hline
    $DO_{t+i} \in \mathbb{R}^{N{\times}K}$ & the complete DO matrix {\color{white}.......}($i=-n+1,...,m$)\\
    \hline
    \end{tabular}
  }
  \label{tab:Notations}
\end{table}

{\bf{3) Compact OD/DO Matrix Generation:}} At each time interval $t$, we measure the numbers of various types of transactions to generate the following vector and matrices.

{\bf{i)}} Incomplete OD Matrix $IOD_t \in \mathbb{R}^{N \times K}$: Specifically, $IOD_t(i,j)$ is the number of passengers that entered station $i$ at time interval $t$ and have exited from station $M_{od}(i,j)$, where $j=1,...,K-1$. Moreover, $IOD_t(i,K)$ is the total number of passengers which entered station $i$ and have exited from the remaining stations.

{\bf{ii)}} Unfinished Order Vector $U_t \in \mathbb{R}^{N}$: This vector is utilized to record the information of ongoing transactions. Specifically, $U_t(i,j)$ denotes the number of passengers that entered at station $i$ at time interval $t$ but have not reached their destinations. Such information has never been explored in previous works \cite{gong2018network,gong2020online,noursalehi2021dynamic}.

{\bf{iii)}} Complete OD Matrix $OD_t \in \mathbb{R}^{N \times K}$: Different to $IOD_t$, this matrix records the complete OD ridership, and it is usually served as the predicted target in our work. Specifically, $OD_t(i,j)$ is the number of passengers that entered station $i$ at time interval $t$ and would exit from station $M_{od}(i,j)$,  where $j=1,...,K-1$. The number of passengers that would head for the remaining stations is stored as $OD_t(i,K)$.

{\bf{iv)}} DO Matrix $DO_t \in \mathbb{R}^{N \times K}$: This matrix denotes the complete DO ridership at time interval $t$. More specifically, $DO_t(i,j)$ is the number of passengers that entered station $M_{do}(i,j)$ at earlier moments and exit from station $i$ at time interval $t$. Similarly, $DO_t(i,K)$ is the total number of exited passengers at the remaining stations.

\begin{myDef} {\bf{Online Metro Origin-Destination Prediction}}
Assuming that the current time interval is $t$ and we aim to utilize the historical ridership of previous $n$ time intervals to forecast the complete OD and DO ridership of future $m$ time intervals:
\begin{equation}
\{IOD, U, DO\}_{t-n+i} {~~}\Rightarrow{~~} \{OD, DO\}_{t+j}
\end{equation}%
where $i=1,2,...,n$ and $j=1,2,...,m$.
\end{myDef}

\section{Methodology}\label{sec:method}
\begin{figure}[t]
    \centering
    \includegraphics[width=0.95\columnwidth]{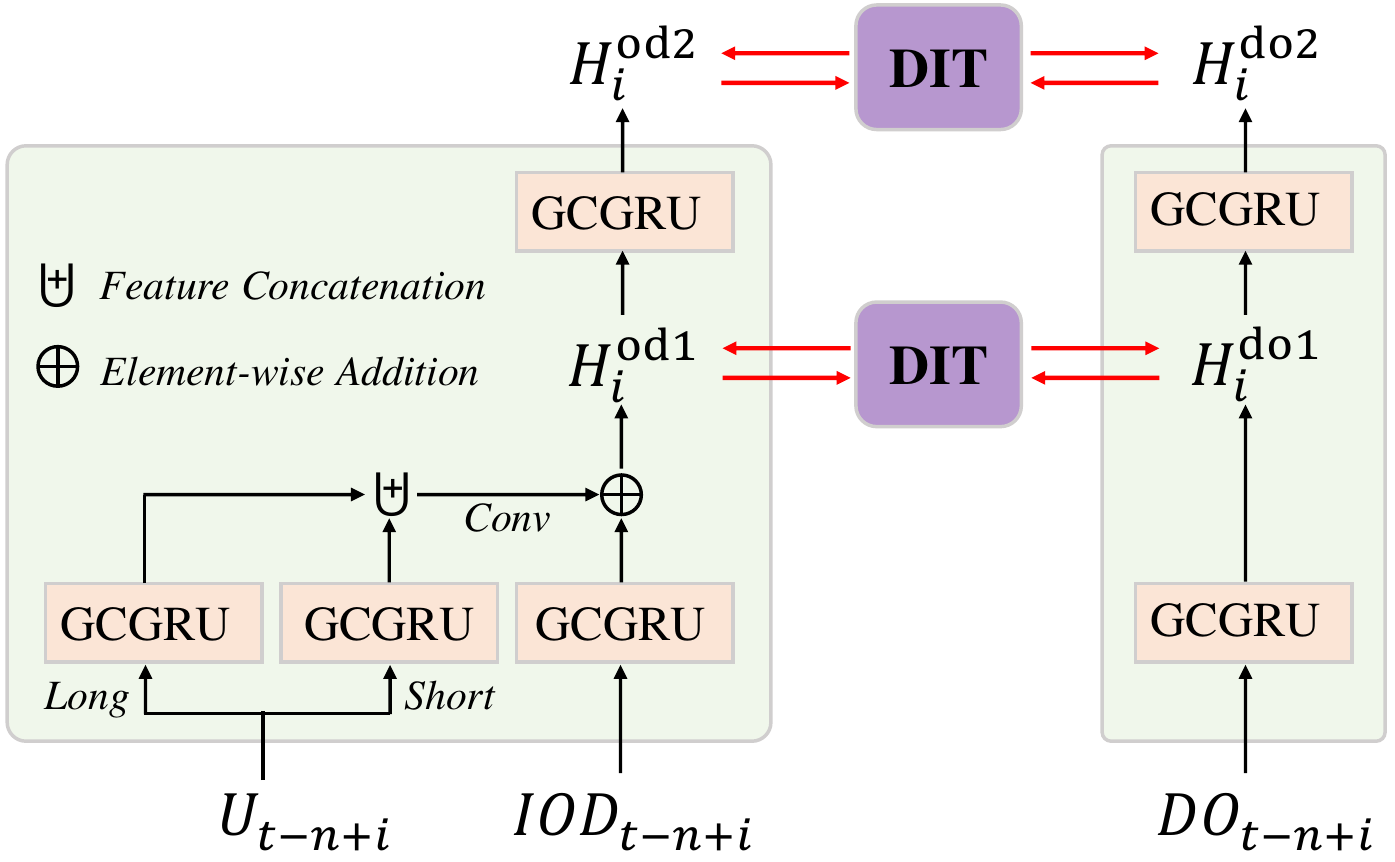}
    \vspace{-3mm}
    \caption{The architecture of the proposed Heterogeneous Information Aggregation Machine. This module is composed of two parallel branches respectively for OD and DO modeling, and a Dual Information Transformer for OD-DO interaction modeling.}
    \label{fig:HIAM}
\end{figure}

In this work, we propose a unified neural network module termed Heterogeneous Information Aggregation Machine (HIAM), which fully captures various information (i.e., incomplete OD ridership, unfinished orders, DO ridership) to effectively model the metro origin-destination distribution. As shown in Fig. \ref{fig:HIAM}, our HIAM consists of an OD modeling branch, a DO modeling branch, and a Dual Information Transformer for OD-DO interaction modeling. Specifically, at each iteration, the OD branch feeds an incomplete OD matrix generated from finished orders and two estimated destination matrices for those unfinished orders into graph convolutional gated recurrent units to learn a compact OD hidden state for memorizing the OD evolution patterns. Meanwhile, the DO branch takes the corresponding DO matrix as input to generate a DO hidden state that memorizes the DO evolution patterns. The Dual Information Transformer is then applied to enhance the OD state and DO state by exploring their mutual information. Based on the proposed HIAM, we develop an online metro origin-destination prediction framework to jointly forecast the complete OD and DO ridership of the next $m$ time intervals.

\subsection{Metro Topology Modeling}
Metro is essentially a complex traffic system with a non-Euclidean topology. Recently, GCN has been proven to be effective for non-Euclidean data embedding \cite{bai2020learning,chen2020knowledge,lin2021learning}. Inspired by these works, we also model a metro system as a directed graph and incorporate it into graph convolution gated recurrent units (GCGRU) to learn spatial-temporal representation for metro ridership.

\begin{figure}[t]
    \centering
    \includegraphics[width=1\columnwidth]{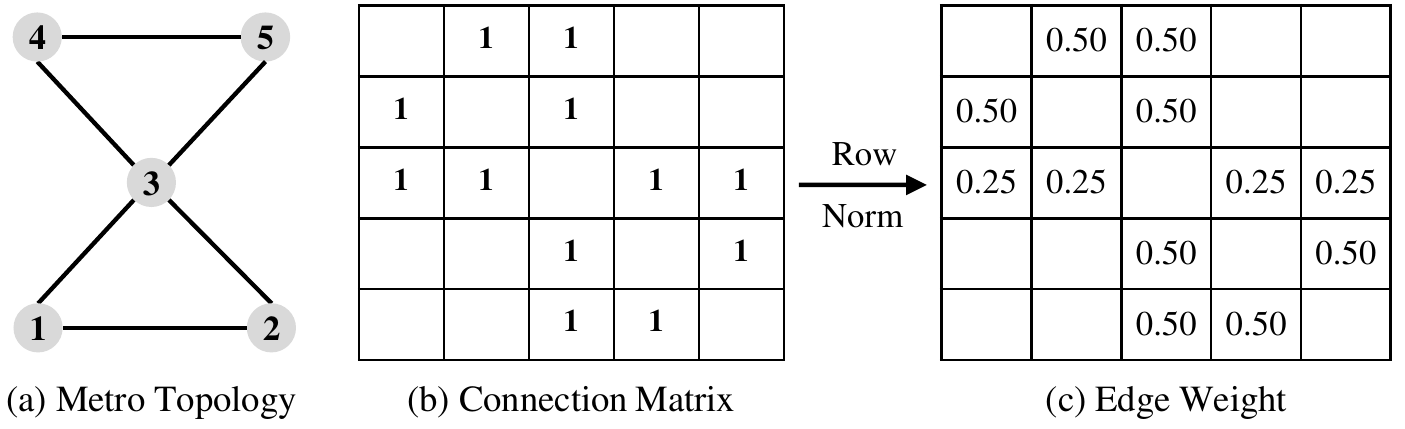}
    \vspace{-7mm}
    \caption{Illustration of the graph construction for non-Euclidean metro systems. (a) is the physical topology of a metro system with five stations. Matrix (b) recodes the connectivity of edges in the constructed graph, while matrix (c) is the normalized weights of edges.}
    \label{fig:metro_graph}
\end{figure}

In this work, we directly utilize the physical topology of the studied metro system to guide the construction of the graph network. By definition, a graph is composed of nodes, edges as well as the weights of edges, i.e., ${G}=(V, E, W)$. Specifically, $V$ is the set of $N$ nodes and each node represents a metro station in the real world. $E \in \mathbb{R}^{N{\times}N}$ is an edge connection matrix. $E(i,j)$ is 1 if the stations $i$ and $j$ are directly connected in the metro system, otherwise, it's set to 0. We then obtain the edge weight matrix $W \in \mathbb{R}^{N{\times}N}$ by applying a linear normalization on each row of $W$:
\begin{equation}
    W(i,j) = \frac{E(i,j)}{\sum_{k=1}^N E(i,k)}.
\end{equation}
One example of graph construction for a metro system with five stations is illustrated in Fig. \ref{fig:metro_graph}.

Graph convolution and gated recurrent units are then integrated for representation learning. Let us assume that the input data of nodes is $I_t = \{I_t^1,I_t^2,...,I_t^N\}$, where $I_t^i$ can be the ridership data $IOD_t(i)$, $U_t(i)$, $DO_t(i)$ or their features. Instead of using spectral convolution \cite{bruna2014spectral,defferrard2016convolutional}, here we utilize spatial convolution \cite{velivckovic2017graph} to aggregate information from neighbor nodes. Specifically, the convolutional feature $f(I_t^i) \in \mathbb{R}^d$ of the node $i$ is computed by:
\begin{equation}
    f(I_t^i)=\Theta_{l}I_t^i + \sum_{j\in \mathcal{N}(i)} W(i,j) \odot \Theta_{n}I_t^j,
\label{eq:gc}
\end{equation}
where $\odot$ is the Hadamard product and $\mathcal{N}_p(i)$ is the neighbor set of node $i$ in the constructed graph. $\Theta_{l}$ denotes the self-loop parameters and $\Theta_{n}$ represents the neighbor parameters. $d$ is the dimension of features.

We then embed graph convolutions into gated recurrent units for temporal modeling. Notice that all original regular convolutions in GRU are replaced with our graph convolutions, and we apply the modified formulation to compute the reset gate, update gate, and new information. For convenience, the computation of the output hidden state $H_t = \{H_t^1,H_t^2,...,H_t^N\}$ is denoted as:
{\begin{equation}
    H_t = \text{GCGRU}(I_t, H_{t-1}), \\
\label{eq:gc_gru_short}
\end{equation}}%
where $H_t^i$ is the hidden state of the $i$-th node and its feature dimension is also set to $d$. Thanks to this graph convolutional gated recurrent unit, we can learn spatial-temporal features effectively from the ridership data of non-Euclidean metro systems.

\subsection{OD Modeling Branch}
In HIAM, an OD modeling branch is specially developed to learn the OD distribution of ridership by jointly exploiting the information of incomplete OD matrices and unfinished order vectors with multiple series-parallel GCGRU. The overall architecture of our OD branch is shown in the left sub-graph of Fig. \ref{fig:HIAM}.

\begin{figure}[t]
    \centering
    \includegraphics[width=1\columnwidth]{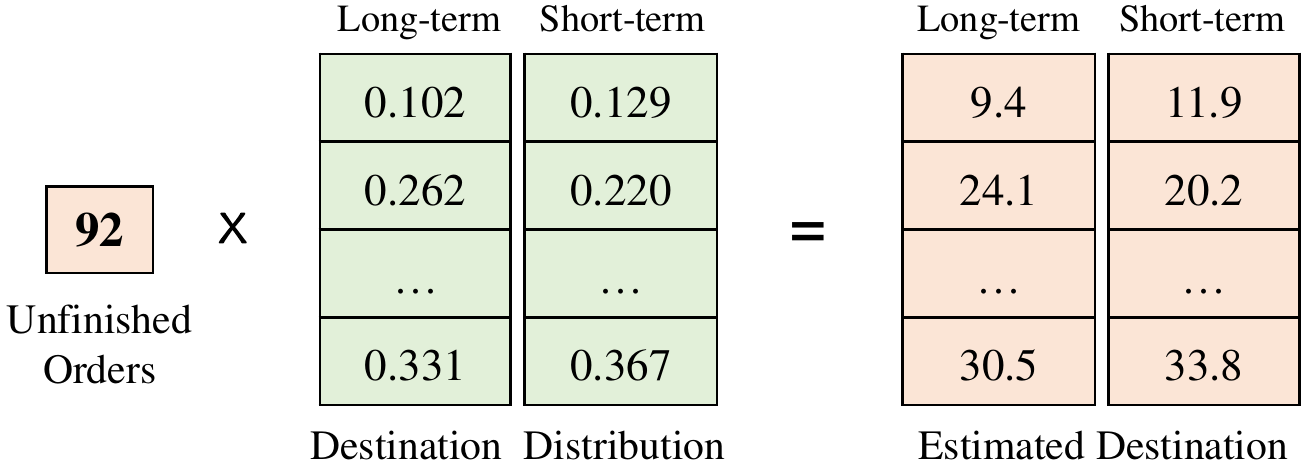}
    \vspace{-7mm}
    \caption{Illustration of the potential destination matrices estimated for unfinished orders. Specifically, we first measure two long short-term destination distributions of historical unfinished orders and then estimate the potential destinations of ongoing passengers.}
    \label{fig:destination_distribution}
\end{figure}

In this work, we fully explore the potential information of unfinished order vectors rather than directly feed them into GCGRU. In particular, considering the periodicity of resident mobilities, we estimate two potential destination matrices for unfinished orders based on their long short-term historical distribution. Let's assume that the current time interval is $t$ and we take the destination estimation for $U_{t-n+i}$ as an example to explain our working mechanism. As shown in Fig. \ref{fig:destination_distribution}, two destination distributions of historical unfinished orders are measured:
\begin{itemize}
\item {\bf{Short-term Destination Distribution}} $DD_{t-n+i}^s\in\mathbb{R}^{N{\times}K}$:
The unfinished order distribution at the same time interval of yesterday is utilized to estimate the destinations of ongoing passengers at the recent time interval $t-n+i$. Specifically, $DD_{t-n+i}^s(j,k)$ is the percentage of ongoing passengers that entered station $j$ at time interval $t-n+i$ of yesterday but have not reached their destination station $M_{od}(j,k)$ until the time interval $t$ of yesterday.

\item {\bf{Long-term Destination Distribution}} $DD_{t-n+i}^l \in \mathbb{R}^{N{\times}K}$:
The unfinished order distribution at the same time interval of the same weekday/weekend is utilized to estimate the destinations of ongoing passengers. Let's assume that today is Monday. $DD_{t-n+i}^s(j,k)$ is the overall percentage of passengers that entered station $j$ at time interval $t-n+i$ of all Monday in the training set but have not reached their destination station $M_{od}(j,k)$ after $n-i$ time intervals.
\end{itemize}

Notice that $DD_{t-n+i}^s$ may be unstable on Monday and Saturday, since the ridership distribution is different between weekdays and weekends. Besides, $DD_{t-n+i}^l$ may be smooth and cannot well reflect the recent ridership distribution. Therefore, both the long short-term distributions are used for destination estimation. Specifically, we generate two potential destination matrices $UOD_{t-n+i}^l$ and $UOD_{t-n+i}^s \in \mathbb{R}^{N{\times}K}$ with following formulations:
\begin{equation}
\begin{split}
    UOD_{t-n+i}^l(j,k) &= U_{t-n+i}(j) * DD_{t-n+i}^l(j,k), \\
    UOD_{t-n+i}^s(j,k) &= U_{t-n+i}(j) * DD_{t-n+i}^s(j,k).
\end{split}
\end{equation}%
These estimated matrices and the corresponding incomplete OD matrix $IOD_{t-n+i}$ are fed into three individual GCGRU for hidden state generation:
\begin{equation}
\begin{split}
    H_i^l &=  GCGRU(UOD_{t-n+i}^l, ~H_{i-1}^l), \\
    H_i^s &=  GCGRU(UOD_{t-n+i}^s, ~H_{i-1}^s), \\
    H_i^{iod} &=  GCGRU(~IOD_{t-n+i}, ~H_{i-1}^{iod}).
\end{split}
\end{equation}%
The hidden states $H_i^l$ and $H_i^s$ are then fused on each node with a 1*1 convolutional layer, whose output is further utilized to complement the hidden state $H_i^{iod}$ of $IOD_{t-n+i}$ and generate a compact OD hidden state $H_i^{od1} \in \mathbb{R}^{N{\times}d}$. This process can be formulated as:
\begin{equation}
    H_i^{od1} = H_i^{iod} + Conv( H_i^l \uplus H_i^s, W_{1*1}), \\
\end{equation}%
where $\uplus$ denotes an operator of feature concatenation and $W_{1*1}$ is the parameters of the convolutional layer.

As shown in Fig. \ref{fig:HIAM}, we then enhance the compact OD state $H_i^{od1}$ with a Dual Information Transformer (DIT) described in Section \ref{sec:DIT}. The enhanced state $\hat{H}_i^{od1}$ is further fed into the following GCGRU and DIT to learn high-order spatial-temporal features. This process can be formulated as:
{\vspace{0mm}
\begin{equation}
    \hat{H}_i^{od1}, \hat{H}_i^{do1} = DIT(H_i^{od1}, ~H_i^{do1}),
    \label{eq:DIT1}
\end{equation}}
{\vspace{-5mm}
\begin{equation}
   ~~ H_i^{od2} =  GCGRU(\hat{H}_i^{od1}, ~\hat{H}_{i-1}^{od2}),
\end{equation}}%
{\vspace{-5mm}
\begin{equation}
    \hat{H}_i^{od2}, \hat{H}_i^{do2} = DIT(H_i^{od2}, ~H_i^{do2}),
    \label{eq:DIT2}
\end{equation}}%
where $H_i^{do1}$ and $H_i^{do2}$ are the hidden states of $DO_{t-n+i}$, and their generation is descried in Section \ref{sec:do_model}.

\subsection{DO Modeling Branch}\label{sec:do_model}
In HIAM, a DO modeling branch is also developed to learn the DO distribution of metro ridership. As shown in the right sub-graph of Fig. \ref{fig:HIAM}, our DO branch is composed of two stacked GCGRU, each of which is followed by a Dual Information Transformer. Specifically, at iteration $i$, the first GCGRU takes the DO matrix $DO_{t-n+i}$ as input to generate an initial DO hidden state $H_i^{do1} \in \mathbb{R}^{N{\times}d}$:
\begin{equation}
    H_i^{do1} =  GCGRU(DO_{t-n+i}, ~H_{i-1}^{do1}).
\end{equation}%
We then employ Eq. (\ref{eq:DIT1}) to enhance this DO state by absorbing the OD information of $H_i^{od1}$. The enhance state $\hat{H}_i^{do1}$ is fed into the second GCGRU and the output hidden state is computed by:
\begin{equation}
   H_i^{do2} = GCGRU(\hat{H}_i^{do1}, \hat{H}_{i-1}^{do2}).
\end{equation}
Similar to the OD branch, $H_i^{do2} \in \mathbb{R}^{N{\times}d}$ is further refined by another Dual Information Transformer, which is formulated as Eq. (\ref{eq:DIT2}). Thanks to the tailor-designed OD-DO interactive mechanism, our method can effectively learn the DO evolutionary trend with the aid of previous OD information.

\subsection{Dual Information Transformer}\label{sec:DIT}
In previous works \cite{gong2018network,gong2020online}, OD and DO ridership are modeled separately. However, we notice that they usually have strong causality and correlation. For OD-to-DO causality, the historical OD ridership essentially affects the future DO ridership since the latter is the spatial-temporal evolution result of the former. Moreover, we can also infer the future OD ridership based on the DO-to-OD correlation. For example, the DO and OD ridership of some tide stations are usually negatively correlated \cite{gong2008data}, e.g., their future OD ridership would increase/decrease when the recent DO ridership decreases/increases.
Taking the causality and correlation into consideration, we propose a novel Dual Information Transformer (DIT) to model the OD and DO distribution jointly by propagating their mutual information in a dual manner. After the interaction, our OD and DO features become more informative for ridership prediction.

\begin{figure}[t]
    \centering
    \includegraphics[width=1\columnwidth]{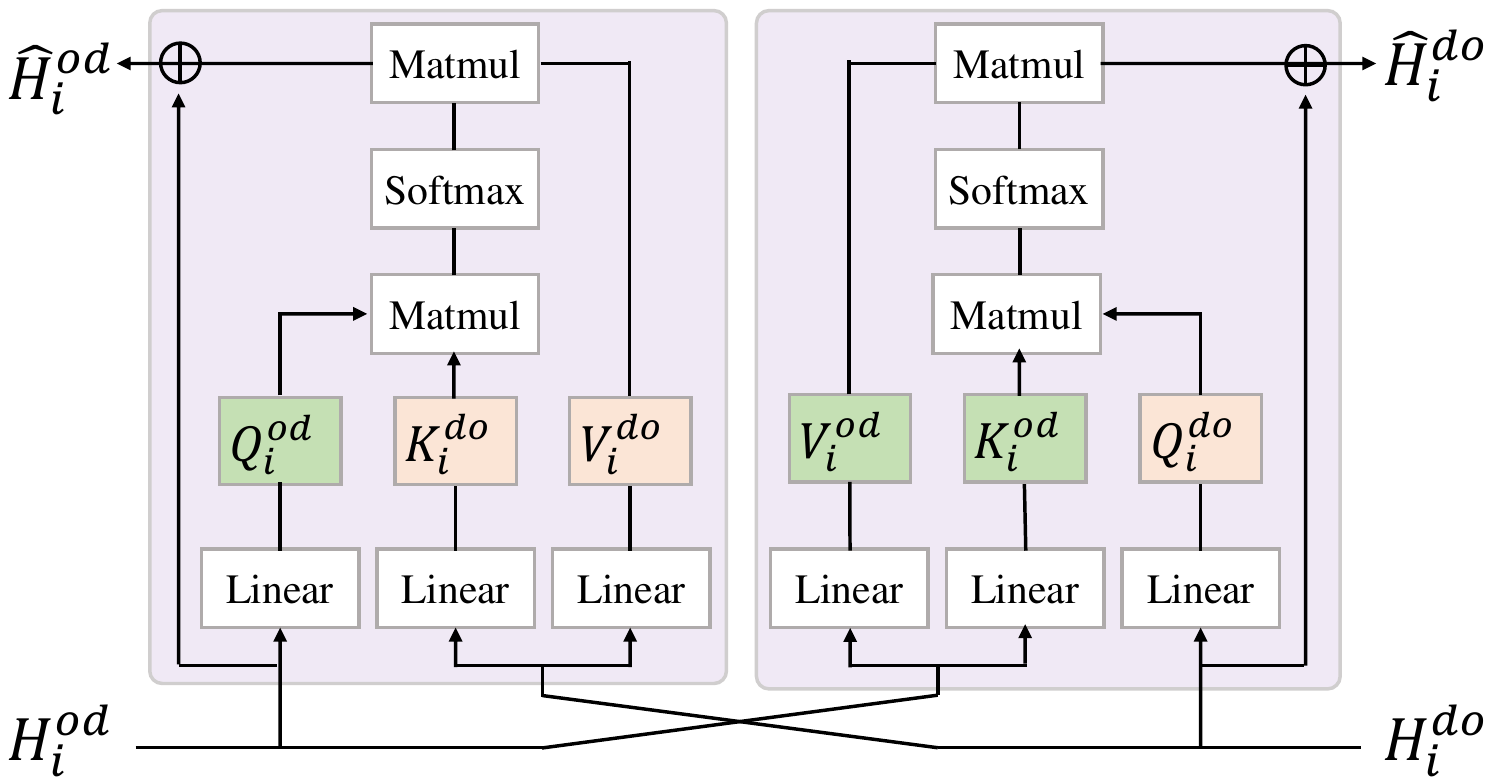}
    \vspace{-6mm}
    \caption{The architecture of the proposed Dual Information Transformer. In this module, we perform information prorogation among OD branch and DO branch to jointly model the OD and DO distribution. The enhanced OD and DO features are more informative for ridership prediction.}
    \label{fig:DIT}
\end{figure}

As shown in Fig. \ref{fig:DIT}, our DIT is implemented with two cross-branch transformers \cite{vaswani2017attention}, where the right one propagates information from OD branch to DO branch, and the left one transfers information in the opposite direction. Here we take as an example the interaction between the OD feature $H_i^{od}$ and DO feature $H_i^{do}$ to illustrate the working mechanism of our DIT. Specifically, $H_i^{od}$ and $H_i^{do}$ are first respectively fed into three linear layers for query, key, and value embedding:
\begin{equation}
\begin{split}
Q_i^{od} &= Conv(H_i^{od}, W^{qod}_{1*1}), {~~}Q_i^{do} = Conv(H_i^{do}, W^{qdo}_{1*1}), \\
K_i^{od} &= Conv(H_i^{od}, W^{kod}_{1*1}), {~~}K_i^{do} = Conv(H_i^{do}, W^{kdo}_{1*1}), \\
V_i^{od} &= Conv(H_i^{od}, W^{vod}_{1*1}), {~~}V_i^{do} = Conv(H_i^{do}, W^{vdo}_{1*1}),
\end{split}
\label{eq:qkv}
\end{equation}
where all linear layers are implemented by 1*1 convolutional layers with individual parameters. Same as $H_i^{od}$ and $H_i^{do}$, these query/key/value features also have a dimension of $N{\times}d$. Based on an attention mechanism, we compute two propagation coefficients $C_i^{o2d} \in \mathbb{R}^{N{\times}N}$ and $C_i^{d2o} \in \mathbb{R}^{N{\times}N}$, which dynamically determines the amount of information propagated among OD feature and DO feature:
\begin{equation}
\begin{split}
C_i^{o2d} &= softmax(Q_i^{do}(K_i^{od})^T), \\
C_i^{d2o} &= softmax(Q_i^{od}(K_i^{do})^T),
\end{split}
\end{equation}
where $T$ denotes a operator of matrix transposition and the softmax function is applied on each column. More specifically, $C_i^{o2d}(j,k)$ is the weight of information transferred from $H_i^{od}(j)$ to $H_i^{do}(k)$, while $C_i^{d2o}(j,k)$ denotes the weight of information transferred from $H_i^{do}(j)$ to $H_i^{od}(k)$. Finally, the cross-branch information prorogation is performed with the following formulations:
\begin{equation}
\begin{split}
\hat{H}_i^{od} &= H_i^{od} + C_i^{d2o}V_i^{do}, \\
\hat{H}_i^{do} &= H_i^{do} + C_i^{o2d}V_i^{od}.
\end{split}
\label{eq:prorogation}
\end{equation}
After the information interaction, the OD state and DO state not only are enhanced mutually, but also better capture the inherent causality/correlation among OD and DO ridership. For convenience, Eq. \ref{eq:qkv}-\ref{eq:prorogation} are simplified as:
\begin{equation}
  \hat{H}_i^{od}, \hat{H}_i^{do} = DIT(H_i^{od}, ~H_i^{do}).
\end{equation}
As stated in Eq. (\ref{eq:DIT1}) and (\ref{eq:DIT2}), the proposed DIT is applied after each level of GCGRU in HIAM for transferring cross-branch information hierarchically.

\begin{figure}[t]
    \centering
    \includegraphics[width=1\columnwidth]{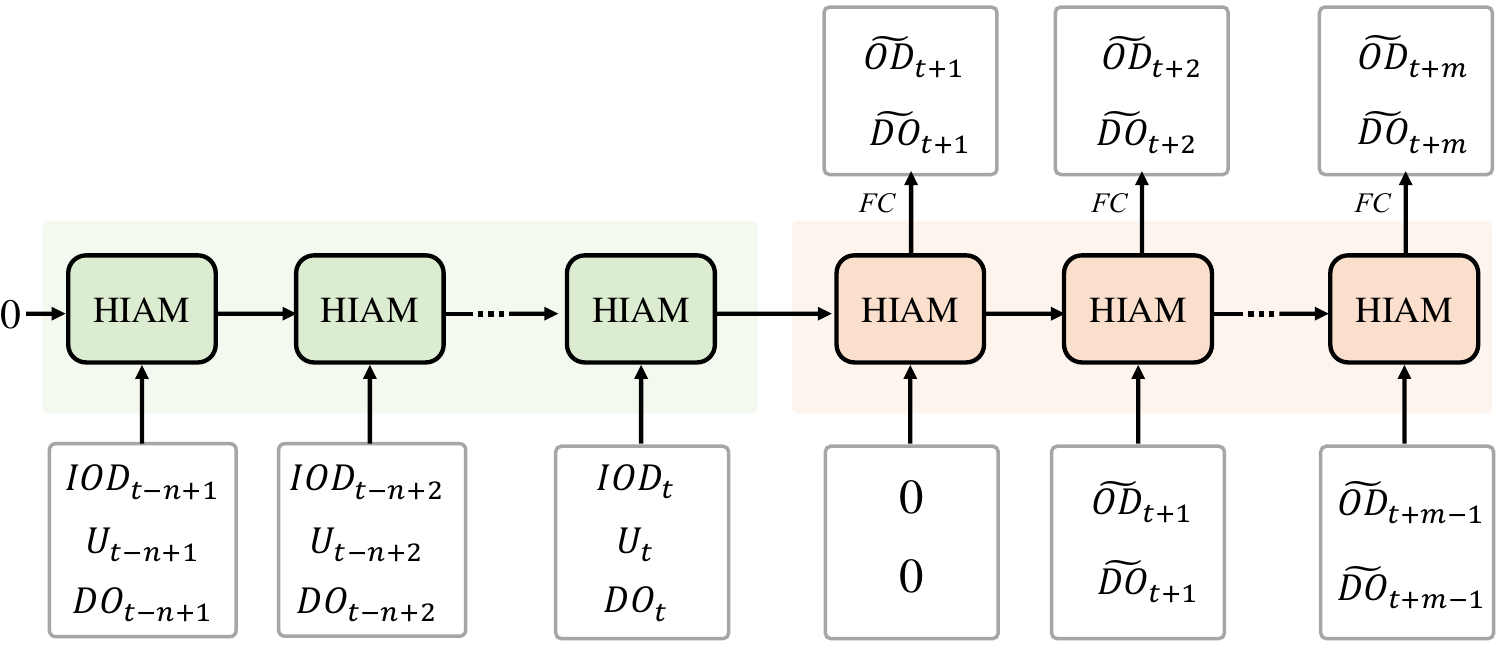}
    \vspace{-6mm}
    \caption{The architecture of our online metro origin-destination prediction framework. The framework is developed with a Seq2Seq architecture, whose encoder and decoder are based on the proposed module, i.e., Heterogeneous Information Aggregation Machine (HIAM).}
    \label{fig:network}
\end{figure}

\subsection{Online Origin-Destination Prediction Framework}
Finally, we apply the proposed HIAM to develop a unified online metro origin-destination prediction framework, which forecasts the OD and DO ridership simultaneously for the next several time intervals. Following previous works \cite{liao2018deep,pan2019urban,zhang2020spatio}, our framework is implemented with a Seq2Seq architecture \cite{sutskever2014sequence}, as shown in Fig. \ref{fig:network}.

Specifically, our framework is composed of an encoder and a decoder, both of which are based on a multi-step HIAM. It is worth noting that the HIAM in the decoder is simplified by removing two GCGRU whose inputs are the estimated destination matrices, because there don't exist unfinished orders at the forecasting stage. Here we introduce the details of our framework.
{\bf{In the encoder}}, at iteration $i$ ($i=1,...,n$), the heterogeneous data $\{IOD_{t-n+i},U_{t-n+i},DO_{t-n+i}\}$ are fed into our HIAM for learning origin-destination evolution information. Notice that the initial hidden states of all GCGRU are set to zero and the final hidden states are used to initialize the hidden states of the decoder.
{\bf{In the decoder}}, the input data of the first iteration is set to zero and the output hidden state of HIAM is respectively fed into a fully connected layers to forecast the OD matrix $\tilde{OD}_{t+1} \in \mathbb{R}^{N{\times}K}$ and DO matrix $\tilde{DO}_{t+1} \in \mathbb{R}^{N{\times}K}$. At iteration $j$ $(j\geq2)$, our HIAM takes as input the predicted ridership matrices of the previous iteration to forecast $\tilde{OD}_{t+j}$ and $\tilde{DO}_{j}$.
After $m$ iterations, we can obtain a sequence of future OD ridership $\{\tilde{OD}_{t+1},..., \tilde{OD}_{t+m}\}$ and a sequence of future OD ridership $\{\tilde{DO}_{t+1},..., \tilde{DO}_{t+m}\}$.

\begin{table}[t]
  \caption{Details of SHMOD and HZMOD datasets. ``\# Stations" denotes the number of metro stations. ``\# OD/DO Pairs'' refers to the number of considered OD/DO pairs
   for matrix compression.}
  \vspace{-2mm}

  \centering
  \resizebox{9cm}{!} {
    \begin{tabular}{c|c|c}
    \hline
    \multicolumn{1}{c|}{\textbf{Dataset}}  & \textbf{SHMOD} & \textbf{HZMOD} \\
    \hline\hline
    City & Shanghai, China & Hangzhou, China \\
    \hline
    \# Stations &  288 & 80 \\
    \hline
    \# OD/DO Pairs & 76 & 26 \\
    \hline
    Daily Ridership & 8.82 M & 2.35 M \\
    \hline
    Time Interval & 15 minutes & 15 minutes \\
    \hline
    {~~~Training Set} & 7/01/2016 - 8/31/2016 & 1/01/2019 - 1/18/2019 \\
    {Validation Set} & 9/01/2016 - 9/09/2016 & 1/19/2019 - 1/20/2019 \\
    {~~~~~Testing Set} & 9/10/2016 - 9/30/2016 & 1/21/2019 - 1/25/2019 \\
    \hline
    \end{tabular}
    }
  \label{tab:datasets}
\end{table}

\section{Experiments}\label{sec:experiment}
In this section, we perform extensive experiments to verify the effectiveness of our model. First, we would introduce the experimental settings (e.g., dataset construction, implementation details, and evaluation strategy). We then compare the proposed method with ten basic and advanced approaches. Finally, we conduct ablation studies to explore the influence of each component in our method.

\subsection{Experiments Settings}
\subsubsection{Dataset Construction}
In this section, we introduce two large-scale benchmarks for online metro origin-destination prediction, which are constructed with billions of transactional records collected from the metro systems of Shanghai and Hangzhou, China. For brevity, these datasets are termed as SHMOD and HZMOD respectively, and their details are summarized in Table \ref{tab:datasets}.

Specifically, SHMOD contains the transactional data of 288 stations, whose period ranges from Jul. 1st, 2016 to Sept. 30th, 2016. HZMOD is built based on the transactional data of 80 stations collected in January 2019. On these datasets, the time interval is set to 15 minutes uniformly. As mentioned above, OD/DO matrices are usually sparse, thus for each station, we mainly consider {\bf {1)}} the number of passengers heading for/coming from its most relevant $K-1$ stations, and {\bf {2)}} the number of passengers heading for/coming from the remaining stations. In this work, $K$ is set to 76 on SHMOD and 26 on HZMOD, because we observe that the ridership of such a small number of OD pairs accounts for a major proportion (i.e., 70\%) of the total ridership. We then employ the data processing method described in Section \ref{sec:preliminary} to generate incomplete OD matrices, unfinished order vectors, and OD matrices for each inferring time interval $t$. Finally, these datasets are officially divided into a training set, a validation set, and a testing set.

\subsubsection{Implementation Details}
In this work, the proposed method is implemented with the PyTorch framework \cite{paszke2019pytorch}. The length $n$ of input sequences is 4, and so is the length $m$ of output sequences. The batch size is set to 8 for SHMOD and 32 for HZMOD, while the feature dimension $d$ is set to 96 uniformly. Xavier uniform \cite{glorot2010understanding} is utilized to initialize the weights of filters and PReLU \cite{he2015delving} is used as the activation function. In our transformer, multi-head attention is adopted and the head number is 4.  During the training phase, the input data and their ground-truths are normalized with Z-score Normalization\footnote{\url{https://en.wikipedia.org/wiki/Standard_score}}. The learning rate is initialized to 0.001 for 60 epochs and decays by 0.2/0.5 for SHMOD/HZMOD every 20 epochs. Adam optimizer \cite{kingma2014adam} is applied to minimize the mean absolute error between the predicted results and the corresponding ground-truths for 300 epochs. All interim models are evaluated on the validation set and the one with the best performance is chosen as our final model, which would be formally evaluated on the testing set for fair comparisons.

\subsubsection{Evaluation Strategy}
As mentioned above, those normalized ground-truths are used as supervision during training. Therefore, when evaluating, we first convert the output OD/DO matrices to the original scale with an inverted Z-score Normalization. Network-wide Mean Absolute Percentage Error (MAPE) is then adopted to evaluate the performance of different methods. In particular, we take into consideration the percentage error of the whole metro network, rather than compute the error of each OD/DO pair separately which is sensitive to a small denominator. Specifically, the MAPE for OD prediction is computed as:
\begin{equation}
MAPE = \frac{\sum_{i=1}^N \sum_{j=1}^{K} |\tilde{OD}(i,j) - OD(i,j)|}{\sum_{i=1}^N \sum_{j=1}^{K} |OD(i,j)|}.
\end{equation}
The MAPE for DO prediction is computed with the same formulation. Notice that our method forecasts the ridership of the next $m$ time intervals, and we would measure the MAPE for each time interval in the following sections.

\begin{table*}[t]
 \caption{Performance of OD prediction and DO prediction on the SHMOD Dataset.}
  \vspace{-2mm}
\newcommand{\tabincell}[2]{\begin{tabular}{@{}#1@{}}#2\end{tabular}}
  \centering
  \resizebox{18.2cm}{!} {
    \begin{tabular}{c|c|c|c|c|c|c|c|c|c|c|c|c}
    \hline
    Ridership & Time Interval	& {\textbf{HA}} & {\textbf{RF}} & {\textbf{LSTM}} & {\textbf{GRU}}& {\textbf{GraphWaveNet}} & {\textbf{DCRNN}} & {\textbf{STG2Seq}} & {\textbf{PVCGN}} & {\textbf{DGSL}} & {\textbf{Informer}} & {\textbf{Ours}} \\
    \hline
    \hline
    \multirow{4}{*}{OD}
     & \textit{15 min }   & 46.28\% & 75.05\% & 43.80\% & 44.96\% & 40.49\% & 41.14\% & 42.56\% & 38.69\% & 40.05\% & 40.06\% & 37.81\%  \\
     & \textit{30 min }   & 46.21\% & 72.19\% & 43.08\% & 42.22\% & 40.20\% & 41.13\% & 41.58\% & 38.69\% & 40.00\% & 40.28\% & 37.79\%  \\
     & \textit{45 min }   & 46.12\% & 71.96\% & 44.55\% & 42.85\% & 40.46\% & 41.55\% & 41.47\% & 38.92\% & 40.32\% & 40.20\% & 37.99\%  \\
     & \textit{60 min }   & 46.05\% & 78.69\% & 46.95\% & 43.99\% & 41.54\% & 42.23\% & 41.77\% & 39.34\% & 40.76\% & 40.48\% & 38.36\%  \\
     \hline
     \multirow{4}{*}{DO}
     & \textit{15 min }   & 47.18\% & 65.47\% & 43.99\% & 42.32\% & 42.29\% & 40.53\% & 40.77\% & 38.72\% & 40.27\% & 40.79\% & 38.65\%  \\
     & \textit{30 min }   & 47.22\% & 65.72\% & 41.91\% & 41.32\% & 42.04\% & 40.59\% & 40.03\% & 39.01\% & 40.54\% & 41.33\% & 38.56\%  \\
     & \textit{45 min }   & 47.23\% & 66.15\% & 42.75\% & 42.16\% & 42.04\% & 41.10\% & 40.22\% & 39.42\% & 41.09\% & 41.71\% & 38.80\%  \\
     & \textit{60 min }   & 47.19\% & 66.75\% & 44.00\% & 43.06\% & 42.18\% & 41.89\% & 40.95\% & 39.96\% & 41.69\% & 42.10\% & 39.18\%  \\
    \hline
    \end{tabular}
  \label{tab:SH_SOTA}
  }
\end{table*}

\begin{table*}[t]
 \caption{Performance of OD prediction and DO prediction on the HZMOD Dataset.}
  \vspace{-2mm}
\newcommand{\tabincell}[2]{\begin{tabular}{@{}#1@{}}#2\end{tabular}}
  \centering
  \resizebox{18.2cm}{!} {
    \begin{tabular}{c|c|c|c|c|c|c|c|c|c|c|c|c}
    \hline
    Ridership & Time Interval	& {\textbf{HA}} & {\textbf{RF}} & {\textbf{LSTM}} & {\textbf{GRU}}& {\textbf{GraphWaveNet}} & {\textbf{DCRNN}} & {\textbf{STG2Seq}}  & {\textbf{PVCGN}}& {\textbf{DGSL}} & {\textbf{Informer}} & {\textbf{Ours}} \\
    \hline
    \hline
    \multirow{4}{*}{OD}
     & \textit{15 min } & 33.00\% & 57.33\% & 32.19\% & 32.85\% & 32.29\% & 31.42\% & 30.54\% & 29.83\% & 30.68\% & 29.66\% &  27.86\%    \\
     & \textit{30 min } & 32.98\% & 55.27\% & 32.46\% & 32.82\% & 32.84\% & 31.61\% & 30.97\% & 30.42\% & 30.87\% & 29.79\% &  27.90\%    \\
     & \textit{45 min } & 32.95\% & 55.88\% & 33.83\% & 34.13\% & 33.52\% & 32.12\% & 31.37\% & 30.84\% & 31.07\% & 30.06\% &  28.04\%  \\
     & \textit{60 min } & 32.91\% & 63.14\% & 35.46\% & 35.90\% & 34.79\% & 32.73\% & 31.46\% & 30.62\% & 31.30\% & 30.60\% &  28.22\%    \\
    \hline
    \multirow{4}{*}{DO }
     & \textit{15 min } & 34.19\% & 52.97\% & 31.23\% & 32.55\% & 33.64\% & 30.89\% & 30.12\% & 30.06\% & 30.58\% & 30.51\% &  28.57\%    \\
     & \textit{30 min } & 34.26\% & 53.11\% & 31.25\% & 31.92\% & 33.76\% & 31.07\% & 29.79\% & 30.49\% & 30.63\% & 30.49\% &  28.64\%   \\
     & \textit{45 min } & 34.31\% & 53.39\% & 32.32\% & 33.14\% & 34.63\% & 31.66\% & 30.40\% & 30.99\% & 31.02\% & 30.90\% &  28.83\%  \\
     & \textit{60 min } & 34.32\% & 53.76\% & 34.02\% & 34.75\% & 35.72\% & 32.46\% & 31.52\% & 31.68\% & 31.57\% & 31.56\% &  29.09\%   \\
    \hline
    \end{tabular}
  \label{tab:HZ_SOTA}
  }
\end{table*}

\subsection{Comparison with State-of-the-Art Methods}\label{sec:SOTA_Comp}
In this work, we compare the proposed method with the following ten basic and advanced approaches. Notice that there are not any public source codes of existing methods for metro OD/DO prediction \cite{zhang2020short,gong2020online,noursalehi2021dynamic}, thus these works are not involved in our comparison.

\begin{itemize}
  \item \textbf{Historical Average (HA):} HA is a periodic baseline that uses historical average ridership to forecast future ridership. More specifically, the OD/DO ridership at 9:00-9:15 am on a specific Monday is predicted as the average of historical observations from the corresponding timestamp of all Mondays in the training set.
  \item \textbf{Random Forest (RF):} As a traditional machine learning method, RF is reimplemented to forecast the future metro ridership with some decision trees. Specifically, the number of trees is set to 10 in our work. These trees are expanded automatically until all leaves contain one sample or are pure.
  \item \textbf{Long Short-Term Memory (LSTM \cite{hochreiter1997long}):} This model employs two fully-connected LSTM layers to forecast the future metro ridership in a Seq2Seq manner \cite{sutskever2014sequence}. The dimension of hidden states is set to 256.
  \item \textbf{Gated Recurrent Unit (GRU \cite{cho2014learning}):} The architecture of this model is similar to that of the previous model. The main difference lies in that this model adopts GRU layers rather than LSTM layers. The dimension of hidden states is also set to 256.
  \item \textbf{GraphWaveNet \cite{wu2019graph}:} In this network, an adaptive dependency matrix is learned to discover graph structure automatically, while a stacked dilated 1D convolution component is developed to model long-range temporal sequences. The official code\footnote{\url{https://github.com/nnzhan/Graph-WaveNet}} is used to reimplement this method on our SHMOD and HZMOD benchmarks.
  \item \textbf{Diffusion Convolutional Recurrent Neural Network (DCRNN \cite{li2018diffusion}):} DCRNN is a representative model for traffic forecasting, in which spatial dependencies are captured through bidirectional random walks on graphs and temporal dependencies are modeled with an encoder-decoder architecture. This model is easily reimplemented to forecast the metro OD/DO ridership with its official code\footnote{\url{https://github.com/liyaguang/DCRNN}}.
  \item \textbf{Spatial-Temporal Graph to Sequence Network (STG2Seq \cite{bai2019stg2seq}):} In this model, spatial and temporal relationships of traffic flow are captured simultaneously with a hierarchical graph convolutional structure. Based on the official code\footnote{\url{https://github.com/LeiBAI/STG2Seq}}, this model is reimplemented for metro OD/DO ridership prediction.
  \item \textbf{Physicla-Virtual Collaborative Graph Network (PVCGN \cite{liu2020physical}):} PVCGN is a recent method designed for metro ridership prediction. In this model, a physical graph, a similarity graph, and a correlation graph are incorporated to learn the complex patterns of metro ridership. The official code\footnote{\url{https://github.com/HCPLab-SYSU/PVCGN}} is adopted to reimplement this model for metro OD/DO prediction.
  \item \textbf{Discrete Graph Structure Learning (DGSL \cite{shang2021discrete}:} This method proposes a probabilistic graph model to learn the graph structure of time series data by optimizing the mean performance over the graph distribution and sampling the discrete graph differentiably. Based on its official code\footnote{\url{https://github.com/chaoshangcs/GTS}}, this method is reimplemented on the SHMOD and HZMOD datasets.
  \item \textbf{Informer \cite{zhou2021informer}:} This is an efficient transformer-based model for long sequence time-series forecasting. This model incorporates three efficient and effective modules to capture long-range dependencies between input and output. Based on its official code\footnote{\url{https://github.com/zhouhaoyi/Informer2020}}, we reimplement Informer to forecast metro OD/DO ridership. Notice that the hyper-parameters input sequence length of the encoder, start token length and prediction sequence length of the decoder are set to 4, 2, and 4, respectively.
\end{itemize}

\begin{table*}[t]
 \caption{Performance of different input information for OD prediction.}
  \vspace{-2mm}
\newcommand{\tabincell}[2]{\begin{tabular}{@{}#1@{}}#2\end{tabular}}
  \centering {
    \begin{tabular}{c|c|c|c|c|c|c}
    \hline
    Dataset & Time Interval & {\textbf{IOD}} & {\textbf{IOD+U}} & {\textbf{IOD+U(short)}}& {\textbf{IOD+U(long)}} & {\textbf{IOD+U(short+long)}} \\
    \hline
    \hline
    \multirow{4}{*}{SHMOD }
     & \textit{15 min }    & 41.04\% & 39.58\% & 39.58\% & 39.38\% & 38.69\%   \\
     & \textit{30 min }  & 41.01\% & 39.61\% & 39.56\% & 39.34\% & 38.43\%       \\
     & \textit{45 min }   & 41.54\% & 40.17\% & 40.08\% & 39.75\% & 38.63\%     \\
     & \textit{60 min }  & 42.43\% & 41.06\% & 40.94\% & 40.43\% & 39.18\%    \\
    \hline
    \multirow{4}{*}{HZMOD}
     & \textit{15 min }   & 31.47\% & 29.89\% & 29.50\% & 29.48\% & 28.75\%     \\
     & \textit{30 min }   & 32.25\% & 30.42\% & 29.88\% & 29.86\% & 28.87\%      \\
     & \textit{45 min }   & 33.60\% & 31.37\% & 30.75\% & 30.55\% & 29.31\%      \\
     & \textit{60 min }   & 35.11\% & 32.60\% & 31.91\% & 31.55\% & 30.05\%        \\
\hline
    \end{tabular}
  \label{tab:unfinished_orders}
  }
\end{table*}

\begin{table*}[t]
 \caption{The influence of OD-to-DO causality on DO prediction performance. Here we explore the OD-to-DO causality to facilitate the metro DO prediction. `\textit{Input} $\rightarrow$ \textit{Output}' denotes that the historical input data is used to forecast the future output data.}
  \vspace{-2mm}
\newcommand{\tabincell}[2]{\begin{tabular}{@{}#1@{}}#2\end{tabular}}
  \centering {
    \begin{tabular}{r|c|c|c|c||c|c|c|c}
    \hline
    \multirow{2}{*}{\textit{Input} $\rightarrow$ \textit{Output}} & \multicolumn{4}{c||}{SHMOD} & \multicolumn{4}{c}{HZMOD} \\
    \cline{2-9}
    & \textit{15 min} & \textit{30 min} & \textit{45 min} & \textit{60 min} & \textit{15 min} & \textit{30 min} & \textit{45 min} & \textit{60 min}  \\
    \hline
    \hline
     IOD $\rightarrow$ DO{~~~~}                  & 41.33\% & 41.26\% & 41.66\% & 42.38\% & 31.81\% & 31.81\% & 32.75\% & 34.12\%  \\
     IOD+U $\rightarrow$ DO{~~~~}                & 40.99\% & 40.81\% & 41.18\% & 41.87\% & 31.57\% & 31.64\% & 32.11\% & 33.05\%  \\
     IOD+U(short+long) $\rightarrow$ DO{~~~~}    & 40.17\% & 39.89\% & 40.05\% & 40.54\% & 30.45\% & 30.31\% & 30.49\% & 31.12\% \\

    \hline
    DO $\rightarrow$ DO{~~~~}                    & 40.41\% & 40.57\% & 41.20\% & 42.05\% & 30.61\% & 31.23\% & 32.44\% & 33.99\% \\
    IOD+U(short+long), DO $\rightarrow$ DO{~~~~} & 38.65\% & 38.56\% & 38.80\% & 39.18\% & 28.57\% & 28.64\% & 28.83\% & 29.09\% \\
    \hline
    \end{tabular}
  }
  \label{tab:causality_res}
\end{table*}

\begin{table*}[t]
 \caption{The influence of DO-to-OD correlation on OD prediction performance. Here the DO-to-OD correlation is incorporated to promote the metro OD prediction. `\textit{Input} $\rightarrow$ \textit{Output}' denotes that the historical input data is used to forecast the future output data.}
  \vspace{-2mm}
\newcommand{\tabincell}[2]{\begin{tabular}{@{}#1@{}}#2\end{tabular}}
  \centering {
    \begin{tabular}{r|c|c|c|c||c|c|c|c}
    \hline
    \multirow{2}{*}{\textit{Input} $\rightarrow$ \textit{Output}} & \multicolumn{4}{c||}{SHMOD} & \multicolumn{4}{c}{HZMOD} \\
    \cline{2-9}
    & \textit{15 min} & \textit{30 min} & \textit{45 min} & \textit{60 min} & \textit{15 min} & \textit{30 min} & \textit{45 min} & \textit{60 min}  \\
    \hline
    \hline
     DO $\rightarrow$ OD{~~~~}                    & 42.31\% & 41.95\% & 42.17\% & 42.87\% & 33.36\% & 33.56\% & 34.28\% & 35.50\% \\
     IOD+U(short+long) $\rightarrow$ OD{~~~~}     & 38.69\% & 38.43\% & 38.63\% & 39.18\% & 28.75\% & 28.87\% & 29.31\% & 30.05\% \\
     IOD+U(short+long), DO $\rightarrow$ OD{~~~~} & 37.81\% & 37.79\% & 37.99\% & 38.36\% & 27.86\% & 27.90\% & 28.04\% & 28.22\% \\
    \hline
    \end{tabular}
  }
  \label{tab:correlation_res}
\end{table*}

The performance of all compared methods are summarized in Table \ref{tab:SH_SOTA} for SHMOD and Table \ref{tab:HZ_SOTA} for HZMOD. We can observe that the traditional method RF obtains unacceptable MAPE at all time intervals and even performs worse than the simple baseline HA, since RF has a limited capacity to capture the complex spatial-temporal distribution of OD/DO ridership. Compared with HA, these RNN-based models LSTM and GRU have certain performance improvements, especially for the first two-step predictions (i.e., 15 and 30 minutes), when explicitly learning the temporal representations from input data. By modeling the spatial/temporal distribution with graph convolutions, those graph-based methods (e.g., GraphWaveNet, STG2Seq, DCRNN, DGSL, and PVCGN) outperform those traditional baselines and common recurrent neural networks. We find that the multi-graph model PVCGN is better than single-graph models in most cases. Moreover, we obverse that the transformer-based model Informer obtains comparable results on the HZMOD dataset, but its performance is not satisfactory on the SHMOD dataset. The reason is that Informer mainly captures temporal dependencies, without learning spatial dependencies explicitly. To our knowledge, the spatial complexity of the Shanghai metro system is much greater than that of the Hangzhou metro system.

Despite progress in model architecture and performance, all the above methods only take finished orders as input data for OD prediction, while ignoring the mutual information between OD and DO distributions. By contrast, our method introduces unfinished orders creatively to enhance the incomplete OD matrices and fully explore the OD-DO information interaction, thereby achieving state-of-the-art performance for both the OD and DO prediction. For instance, on the SHMOD dataset, our HIAM obtains the lowest MAPE 38.36\% and 39.18\% respectively for OD ridership and DO ridership in terms of 60-minute prediction. On the HZMOD dataset, our method also outperforms all previous approaches consistently with large margins during each time interval. These comparisons greatly demonstrate the effectiveness of the proposed HIAM for online metro origin-destination prediction.

\subsection{Ablation Studies}
In this subsection, we perform extensive analyses to verify the effectiveness of each component of the proposed HIAM.

\subsubsection{\textbf{Effectiveness of Unfinished Order Information}}
As mentioned above, to facilitate the OD prediction, our method takes unfinished transactions into consideration and estimates the potential destinations of ongoing passengers based on long short-term historical distributions. To show the effectiveness of our unfinished order usage strategy, we implement five variants of our method for OD modeling. Note that these variants don't involve the DO prediction.
\begin{itemize}
\item \textbf{IOD-Net:} This variant directly utilizes these previous incomplete OD matrices $\{IOD_{t-n+i} | i=1,...,n\}$ to forecast the future complete OD ridership.
\item \textbf{IOD+U-Net:} This variant takes these incomplete OD matrices and those corresponding unfinished order vectors $\{U_{t-n+i} | i=1,...,n\}$ for OD prediction. Note that these unfinished order vectors are directly fed into GCGRU without potential destination estimation.
\item \textbf{IOD+U(short)-Net:} The architecture of this variant is similar to that of IOD+U-Net. In this network, the potential destinations of each $U_{t-n+i}$ are first estimated on the basis of the short-term destination distribution of historical unfinished orders.
\item \textbf{IOD+U(long)-Net:} Different from the previous variant, this network employs the long-term destination distribution of historical unfinished orders to estimate the potential destinations of those ongoing passengers.
\item \textbf{IOD+U(short+long)-Net:} This network incorporates both long-term and short-term historical distributions to estimate the passengers' potential destinations for enhancing the incomplete OD information.
\end{itemize}

The performance of all variants is summarized in Table \ref{tab:unfinished_orders}. We can observe that IOD-Net obtains poor MAPE at all time intervals on both datasets, since this model uses very limited information for ridership prediction. By directly introducing unfinished data, IOD+U-Net can decrease the MAPE from 42.43\% to
41.06\% on SHMOD and from 35.11\% to 32.60\% on HZMOD for the 60-minute prediction. This phenomenon well demonstrates the significance of the information aggregation from unfinished orders. Moreover, the variants IOD+U(short)-Net and IOD+U(long)-Net perform better than IOD+U-Net, when estimating the potential destinations of unfinished orders explicitly. For instance, at the fourth time interval, IOD+U(short)-Net obtains a MAPE 31.91\% on HZMOD, while IOD+U(long)-Net obtains a MAPE 40.43\% on SHMOD. Finally, IOD+U(short+long)-Net achieves obvious performance improvement by incorporating the long short-term estimated destination information. Specifically, compared with IOD-Net, this model reduces the MAPE, on average, by 2.77\% on the SHMOD dataset and by 3.86\% on the HZMOD dataset. Therefore, we conclude that the usage of unfinished orders with long short-term destination estimation can greatly promote online OD prediction.

\subsubsection{\textbf{Influence of OD-to-DO Causality}}
In this subsection, we explore the influence of OD-to-DO causality on metro DO prediction, i.e., utilizing the historical OD ridership information to forecast the future DO ridership. Here we implement the following variants that do not use historical DO information but purely exploit OD-to-DO causality for DO prediction.
\begin{itemize}
\item \textbf{IOD$\rightarrow$DO:} This variant only takes incomplete OD matrices $\{IOD_{t-n+i} | i=1,...,n\}$ to forecast the future DO ridership $\{DO_{t+j} | i=1,...,m\}$.
\item \textbf{IOD+U$\rightarrow$DO:} This variant forecasts the future DO ridership from historical $\{IOD_{t-n+i}, U_{t-n+i} | i=1,...,n\}$. Notice that the potential destinations of $U_{t-n+i}$ are not estimated in this variant.
\item \textbf{IOD+U(short+long)$\rightarrow$DO:} In this variant, historical incomplete OD matrices and the long short-term estimated destinations of unfinished orders are incorporated to predict the future DO ridership.
\end{itemize}

The performance of the above variants is summarized in Table \ref{tab:causality_res}. Specifically, the IOD$\rightarrow$DO model performs poorly, since incomplete OD matrices only record the number of passengers who have completed their trips. We can observe that the performance of those OD-to-DO models can be further improved when incorporating the information about unfinished orders. In particular, the IOD+U(short+long)$\rightarrow$DO model achieves very competitive performance, even better than the DO$\rightarrow$DO model that uses the historical DO ridership to forecast the future DO ridership. Finally, as shown in the last row of Table \ref{tab:causality_res}, the IOD+U(short+long),DO$\rightarrow$DO model achieves the best performance on both datasets, when learning the OD-to-DO causality and DO-to-DO mapping simultaneously. These experiments demonstrate that OD-to-DO causality can well facilitate the metro DO ridership prediction.

\subsubsection{\textbf{Influence of DO-to-OD Correlation}}
In \cite{noursalehi2021dynamic}, only the historical DO ridership was used to forecast the future OD ridership, due to the delayed availability of complete OD matrices. In this subsection, we delve into the impact of DO-to-OD correlation on OD prediction. Here we implement a variant of our model termed DO$\rightarrow$OD, which utilizes $\{DO_{t-n+i} | i=1,...,n\}$ to forecast $\{OD_{t+j} | j=1,...,m\}$. As shown in Table \ref{tab:correlation_res}, the DO$\rightarrow$OD variant obtains unsatisfactory MAPE on both datasets, performing much worse than the IOD+U(short+long)$\rightarrow$OD variant that exploits incomplete OD matrices and unfinished orders for future OD prediction. However, we find that the performance of OD prediction can be further improved, when the DO-to-OD correlation and IOD+U(short+long)-to-OD mapping are learned simultaneously. Specifically, as shown in the last row of Table \ref{tab:correlation_res}, the IOD+U(short+long),DO$\rightarrow$OD variant reduces the MAPE, on average, by 0.75\% on the SHMOD dataset and by 1.24\% on the HZMOD dataset, compared with the IOD+U(short+long)$\rightarrow$OD variant. Therefore, we can draw the following conclusions. {\textbf{i)}} It is inappropriate to perform OD prediction only using the historical DO ridership. {\textbf{ii)}} The DO-to-OD correlation can facilitate the metro OD ridership prediction to a certain extent.

\begin{table}[t]
 \caption{Performance of different OD-DO interaction methods on the SHMOD and HZMOD datasets.}
  \vspace{-2mm}
\newcommand{\tabincell}[2]{\begin{tabular}{@{}#1@{}}#2\end{tabular}}
  \centering
   \resizebox{9cm}{!} {
    \begin{tabular}{c|c|c|c|c|c}
    \hline
    Dataset & Ridership & Time Interval & {\textbf{W/O}}	& {\textbf{SS}} & DIT \\
    \hline
    \hline
    \multirow{8}{*}{SHMOD} & \multirow{4}{*}{OD}
     &   \textit{15 min }   & 38.69\% & 38.66\% & 37.81\%      \\
     & & \textit{30 min } & 38.43\% & 38.57\% & 37.79\%       \\
     & & \textit{45 min }  & 38.63\% & 38.84\% & 37.99\%     \\
     & & \textit{60 min } & 39.18\% & 39.24\% & 38.36\%    \\
     \cline{2-6}
     & \multirow{4}{*}{DO}
     &   \textit{15 min }   & 40.41\% & 39.30\% & 38.65\%     \\
     & & \textit{30 min } & 40.57\% & 39.22\% & 38.56\%      \\
     & & \textit{45 min }  & 41.20\% & 39.51\% & 38.80\%     \\
     & & \textit{60 min } & 42.05\% & 39.94\% & 39.18\%    \\
    \hline
    \hline
    \multirow{8}{*}{HZMOD} &\multirow{4}{*}{OD}
     &   \textit{15 min }  & 28.75\% & 28.49\% & 27.86\%       \\
     & & \textit{30 min }  & 28.87\% & 28.51\% & 27.90\%      \\
     & & \textit{45 min }  & 29.31\% & 28.74\% & 28.04\%       \\
     & & \textit{60 min }  & 30.05\% & 29.13\% & 28.22\%        \\
     \cline{2-6}
     & \multirow{4}{*}{DO}
     &   \textit{15 min }  & 30.61\% & 29.17\% & 28.57\%       \\
     & & \textit{30 min }  & 31.23\% & 29.14\% & 28.64\%        \\
     & & \textit{45 min }  & 32.44\% & 29.35\% & 28.83\%        \\
     & & \textit{60 min }  & 33.99\% & 29.77\% & 29.09\%         \\
     \hline
    \end{tabular}
  \label{tab:interaction}
  }
\end{table}

\subsubsection{\textbf{Exploration of OD-DO Interaction Operation}}
In this work, we introduce a novel Dual Information Transformer (DIT) to capture the mutual information among OD distribution and DO distribution. Here we explore the influences of different OD-DO interaction operations:
\begin{itemize}
\item {\bf{W/O Interaction}}: This method doesn't perform OD-DO interaction, which means that the future OD ridership and DO ridership are forecasted separately.
\item {\bf{Single-Station (SS) Interaction}}: This method propagates the OD and DO information on the same station. That is to say, the OD information of station $i$ is only used to enhance the DO information of station $i$, and vice versa. More specifically, the OD hidden state and DO hidden state of station $i$ are concatenated and fed into a 1*1 convolutional layer to generate the enhanced OD hidden state and DO hidden state.
\item {\bf{DIT Interaction}}: The method propagates information between all stations with the proposed DIT, in which the OD/DO information of station $i$ can be transferred to enhance the DO/OD information of station $j$.
\end{itemize}

The performance of all OD-DO interaction methods are summarized in Table \ref{tab:interaction}. We can observe that the model with Single-Station Interaction surpasses the model without interaction significantly for DO prediction, and can also improve the performance of OD prediction to a certain extent, since the causality of OD-to-DO is easily captured but the correlation of DO-to-OD is challenging for Single-Station Interaction. When adopting cross-station interaction, our DIT can fully exploit the OD-DO mutual information, thereby achieving the best performance for both OD and DO prediction. For instance, compared with ``W/O Interaction'', our DIT reduces the MAPE, on average, by 0.75\% for OD prediction and 2.26\% for  DO prediction on the SHMOD dataset. Similar performance improvement can be obtained on the HZMOD dataset. These experiments verify the superiority of the proposed DIT for OD-DO information interaction.

\begin{table}[t]
 \caption{Performance of different lengths of the input sequence on the SHMOD dataset.}
  \vspace{-2mm}
\newcommand{\tabincell}[2]{\begin{tabular}{@{}#1@{}}#2\end{tabular}}
  \centering {
  \resizebox{9cm}{!} {
    \begin{tabular}{c|c|c|c|c|c|c}
    \hline
    \multirow{2}{*}{Ridership} & \multirow{2}{*}{Time Interval} &
    \multicolumn{5}{c}{Input Sequence Length} \\
    \cline{3-7}
    &  & {\textbf{1}}	& {\textbf{2}} & {\textbf{3}} & {\textbf{4}}& {\textbf{5}} \\
    \hline
    \hline
    \multirow{4}{*}{OD}
     & \textit{15 min }   & 38.99\% & 38.20\% & 37.90\% & 37.81\% & 37.79\%  \\
     & \textit{30 min } & 39.00\% & 38.24\% & 37.88\% & 37.79\% & 37.72\%  \\
     & \textit{45 min }  & 39.19\% & 38.53\% & 38.12\% & 37.99\% & 37.93\%  \\
     & \textit{60 min } & 39.46\% & 38.94\% & 38.48\% & 38.36\% & 38.39\% \\
    \hline
    \multirow{4}{*}{DO}
     & \textit{15 min }   & 39.24\% & 38.82\% & 38.71\% & 38.65\% & 38.55\% \\
     & \textit{30 min } & 39.06\% & 38.80\% & 38.63\% & 38.56\% & 38.46\%  \\
     & \textit{45 min }  & 39.35\% & 39.10\% & 38.91\% & 38.80\% & 38.68\%  \\
     & \textit{60 min } &39.74\% & 39.54\% & 39.31\% & 39.18\% & 39.04\% \\
    \hline
    \end{tabular}
  }
  \label{tab:input_len_SH}
  }
\end{table}

\begin{table}[t]
 \caption{Performance of different lengths of the input sequence on the HZMOD dataset.}
  \vspace{-2mm}
\newcommand{\tabincell}[2]{\begin{tabular}{@{}#1@{}}#2\end{tabular}}
  \centering {
    \resizebox{9cm}{!} {
    \begin{tabular}{c|c|c|c|c|c|c}
    \hline
    \multirow{2}{*}{Ridership} & \multirow{2}{*}{Time Interval} &
    \multicolumn{5}{c}{Input Sequence Length} \\
    \cline{3-7}
    &  & {\textbf{1}}	& {\textbf{2}} & {\textbf{3}} & {\textbf{4}}& {\textbf{5}} \\
    \hline
    \hline
        \multirow{4}{*}{OD}
     & \textit{15 min }  & 28.44\% & 28.02\% & 27.96\% & 27.86\% & 27.82\%      \\
     & \textit{30 min }  & 28.45\% & 28.12\% & 28.00\% & 27.90\% & 27.84\%      \\
     & \textit{45 min }  & 28.58\% & 28.29\% & 28.19\% & 28.04\% & 28.02\%       \\
     & \textit{60 min }  & 28.72\% & 28.52\% & 28.41\% & 28.22\% & 28.26\%       \\
    \hline
    \multirow{4}{*}{DO}
     & \textit{15 min }  & 29.15\% & 28.81\% & 28.58\% & 28.57\% & 28.46\%      \\
     & \textit{30 min }  & 29.16\% & 28.99\% & 28.66\% & 28.64\% & 28.47\%      \\
     & \textit{45 min }  & 29.33\% & 29.24\% & 28.88\% & 28.83\% & 28.64\%     \\
     & \textit{60 min }  & 29.67\% & 29.55\% & 29.18\% & 29.09\% & 28.95\%      \\
     \hline
    \end{tabular}
    }
  \label{tab:input_len_HZ}
  }
\end{table}

\begin{figure}[b]
    \centering
    \includegraphics[width=1\columnwidth]{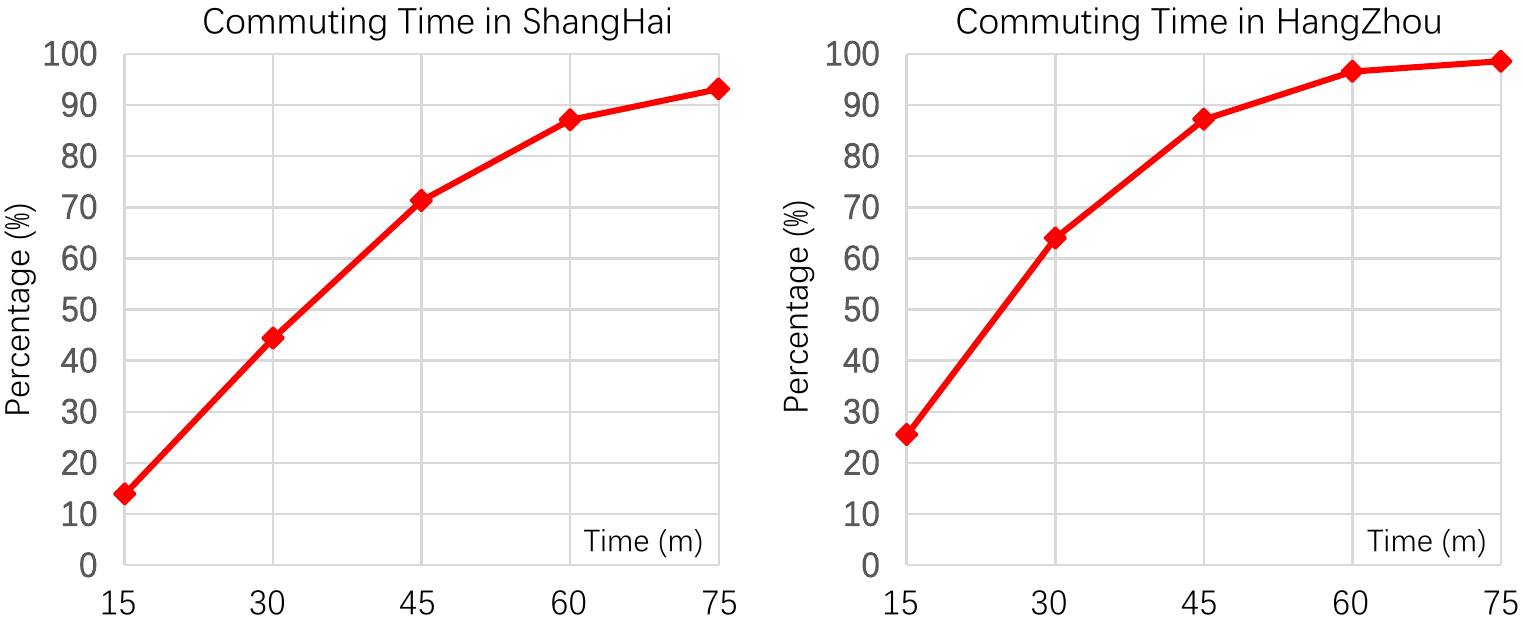}
    \vspace{-6mm}
    \caption{The Cumulative Density Function of the commuting time of metro passengers in Shanghai and Hangzhou. We can observe that the commuting time of most passengers is within one hour.}
    \label{fig:commuting_time}
\end{figure}


\begin{table*}[t]
 \caption{Performance of compressed OD/DO matrices and original OD/DO matrices on the SHMOD dataset.}
  \vspace{-2mm}
\newcommand{\tabincell}[2]{\begin{tabular}{@{}#1@{}}#2\end{tabular}}
  \centering {
    \begin{tabular}{c|c|c|c|c|c}
    \hline
    \multirow{2}{*}{Ridership} & \multirow{2}{*}{Time Interval} &
    \multicolumn{2}{c|}{With Compression}& \multicolumn{2}{c}{Without Compression} \\
     \cline{3-6}
     &  & {Top $K-1$ Stations} & Remaining Stations (Merged) & Top $K-1$ Stations & Remaining Stations (Non-merged) \\
    \hline
    \hline

    \multirow{4}{*}{OD}
     & \textit{15 min } & 47.56\% & 14.14\% & 48.36\% & 84.40\% \\
     & \textit{30 min } & 47.30\% & 14.54\% & 48.04\% & 84.15\% \\
     & \textit{45 min } & 47.40\% & 14.96\% & 48.15\% & 84.10\% \\
     & \textit{60 min } & 47.75\% & 15.36\% & 48.55\% & 84.15\% \\

         \hline
    \multirow{4}{*}{DO}
     & \textit{15 min } & 49.09\% & 14.04\% & 50.05\% & 82.63\% \\
     & \textit{30 min } & 48.91\% & 14.25\% & 49.82\% & 82.56\% \\
     & \textit{45 min } & 49.11\% & 14.62\% & 50.00\% & 82.64\% \\
     & \textit{60 min } & 49.52\% & 14.95\% & 50.41\% & 82.78\% \\
\hline
    \end{tabular}
  \label{tab:all_stations_SHMOD}
  }
\end{table*}

\begin{table*}[t]
 \caption{Performance of compressed OD/DO matrices and original OD/DO matrices on the HZMOD dataset.}
  \vspace{-2mm}
\newcommand{\tabincell}[2]{\begin{tabular}{@{}#1@{}}#2\end{tabular}}
  \centering {
    \begin{tabular}{c|c|c|c|c|c}
    \hline
    \multirow{2}{*}{Dataset} & \multirow{2}{*}{Time Interval} &
    \multicolumn{2}{c|}{With Compression}& \multicolumn{2}{c}{Without Compression} \\
     \cline{3-6}
     &  & {Top $K-1$ Stations} & Remaining Stations (Merged) & Top $K-1$ Stations & Remaining Stations (Non-merged) \\
    \hline
    \hline
    \multirow{4}{*}{OD}
     & \textit{15 min } & 33.56\% & 13.31\% & 34.01\% & 62.42\%  \\
     & \textit{30 min } & 33.57\% & 13.44\% & 33.94\% & 61.89\%  \\
     & \textit{45 min } & 33.71\% & 13.60\% & 34.22\% & 61.89\%  \\
     & \textit{60 min } & 33.94\% & 13.65\% & 34.60\% & 62.21\%  \\
\hline
    \multirow{4}{*}{DO}
     & \textit{15 min } & 34.88\% & 13.32\% & 35.40\% & 61.14\%  \\
     & \textit{30 min } & 34.86\% & 13.59\% & 35.28\% & 60.66\%  \\
     & \textit{45 min } & 35.02\% & 13.85\% & 35.48\% & 60.90\%  \\
     & \textit{60 min } & 35.29\% & 14.05\% & 35.92\% & 61.45\%  \\
\hline
    \end{tabular}
  \label{tab:all_stations_HZMOD}
  }
\end{table*}

\subsubsection{\textbf{Impact of Different Input Sequence Length}}
As described in Section \ref{sec:preliminary}, we utilize the data of previous $n$ time intervals to forecast the ridership of future $m$ time intervals. Following \cite{liu2020physical}, we fix the output sequence length $m$ to 4 and explore the effect of input sequence length $n$ for online origin-destination prediction. As shown in Table \ref{tab:input_len_SH} and Table \ref{tab:input_len_HZ}, the MAPE gradually decreases as the variate $n$ increase from 1 to 4, and longer sequence no longer results in significant improvement. To explain this phenomenon, we measure the commuting time of metro passengers. As shown in Fig. \ref{fig:commuting_time}, we can see that most passengers (i.e., 87.10\% for ShangHai and 96.54\% for HangZhou) complete their metro journey within one hour, i.e., four time intervals. It is reasonable to use the OD/DO ridership of the previous hour to predict the OD/DO ridership of the next hour. Therefore, the length $m$ of the input sequence is set to 4 consistently in this work.

\subsubsection{\textbf{Are compressed OD/DO matrices useful?}}
As mentioned in Section \ref{sec:preliminary}, we compress those original OD/DO matrices with a dimension $N{\times}N$ to form compact matrices with a dimension $N{\times}K$, i.e., forecasting the OD/DO ridership of top $K-1$ most relevant stations and the total ridership of remaining stations. In this subsection, we explore the influence of matrix compression for online metro origin-destination prediction. To this end, we implement a variant of our model that utilizes the historical $N{\times}N$ ridership matrices to forecast the future $N{\times}N$ ridership matrices. Here we measure the prediction performance of the top $K-1$ stations and the remaining stations separately.

As shown in Table \ref{tab:all_stations_SHMOD}, the MAPE of small cross-station ridership is more than 80\% on the SHMOD dataset when we use original sparse matrices, since the ridership between weakly-relevant stations usually lacks regularity. Such poor performance is unacceptable and meaningless for practical application. By contrast, when merging these weakly relevant stations, our method obtains a small MAPE of about 14\% for OD and DO prediction, since the evolution patterns of the total ridership of these stations are easily captured. Moreover, we find that introducing the non-merged ridership between weakly relevant stations would degrade the prediction performance between highly relevant stations. Specifically, the MAPE of the top $K-1$ stations of non-compressed matrices is 49.17\% on average, while that of compressed matrices is 48.33\% on average. This is because the irregular ridership of weakly relevant stations confuses the prediction model to a certain degree. As shown in Table \ref{tab:all_stations_HZMOD}, we can observe that the performance of compressed OD/DO matrices is also better on the HZMOD dataset. In summary, the OD/DO matrix compression is meaningful for online origin-destination prediction.

\section{Conclusion}\label{sec:conclusion}
In this work, we focus on a crucial yet challenging task, online metro origin-destination prediction, i.e., forecasting the OD ridership and DO ridership for multiple time intervals in the future. However, conventional methods either directly used the limited information of incomplete OD matrix for inference, or completely neglected the causality and correlation between these two types of cross-station ridership. To facilitate this problem, we introduce a novel Heterogeneous Information Aggregation Machine (HIAM), which fully exploits heterogeneous information of historical data (e.g., incomplete OD matrices, unfinished order vectors, and DO matrices)  to jointly learn the spatial-temporal patterns of OD and DO ridership. 
%
Based on HIAM, we develop a Seq2Seq network to forecast the future OD and DO ridership simultaneously. Finally, we conduct extensive experiments on two large-scale benchmarks, and experiment results show that our method achieves state-of-the-art performance for online metro origin-destination prediction.

\ifCLASSOPTIONcompsoc
\else
\fi

\ifCLASSOPTIONcaptionsoff
  \newpage
\fi

\normalem
\bibliographystyle{IEEEtran}
\bibliography{sample-base}

\begin{IEEEbiography}[{\includegraphics[width=1in,height=1.25in,clip,keepaspectratio]{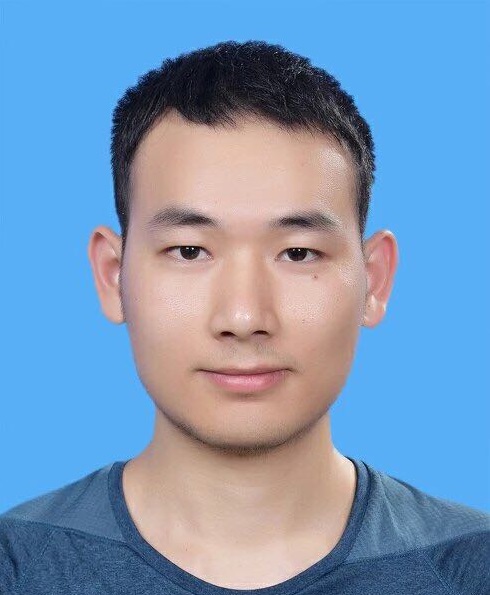}}]{Lingbo Liu}
received the Ph.D degree from the School of Computer Science and Engineering, Sun Yat-sen University, Guangzhou, China, in 2020. From March 2018 to May 2019, he was a research assistant at the University of Sydney, Australia. His current research interests include machine learning and urban computing. He has authorized and co-authorized on more than 20 papers in top-tier academic journals and conferences. He has been serving as a reviewer for numerous academic journals and conferences such as TPAMI, TKDE, TNNLS, TITS, CVPR, ICCV and IJCAJ.
\end{IEEEbiography}

\begin{IEEEbiography}[{\includegraphics[width=1in,height=1.25in,clip,keepaspectratio]{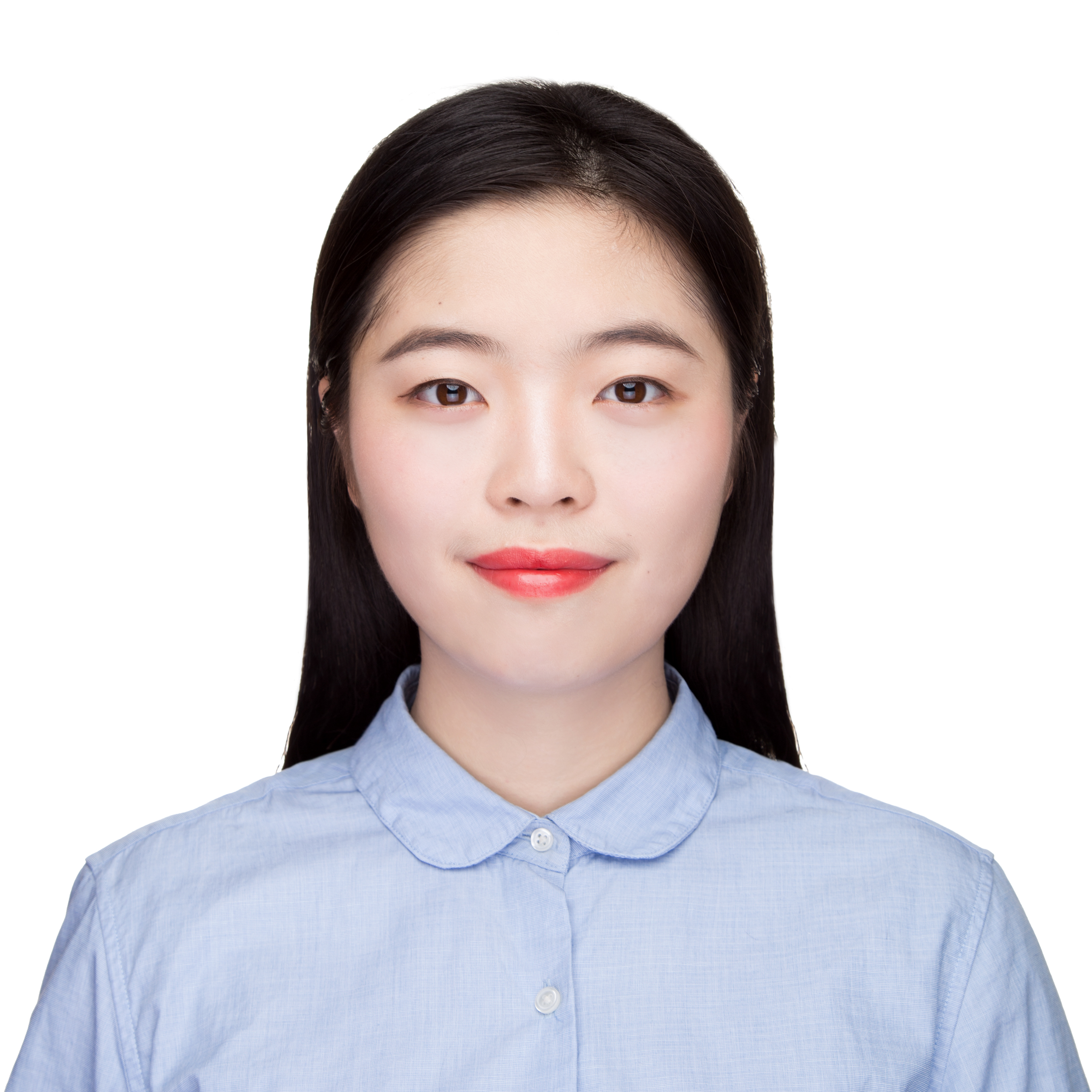}}]
{Yuying Zhu}
 received the B.E. degree from the School of Informatics, Xiamen University, Xiamen, China, in 2020, and she is currently pursuing the Master's degree in computer science in the School of Computer Science and Engineering, Sun Yat-sen University, Guangzhou, China. Her current research interests include deep learning and data mining.
\end{IEEEbiography}

\begin{IEEEbiography}[{\includegraphics[width=1in,height=1.25in,clip,keepaspectratio]{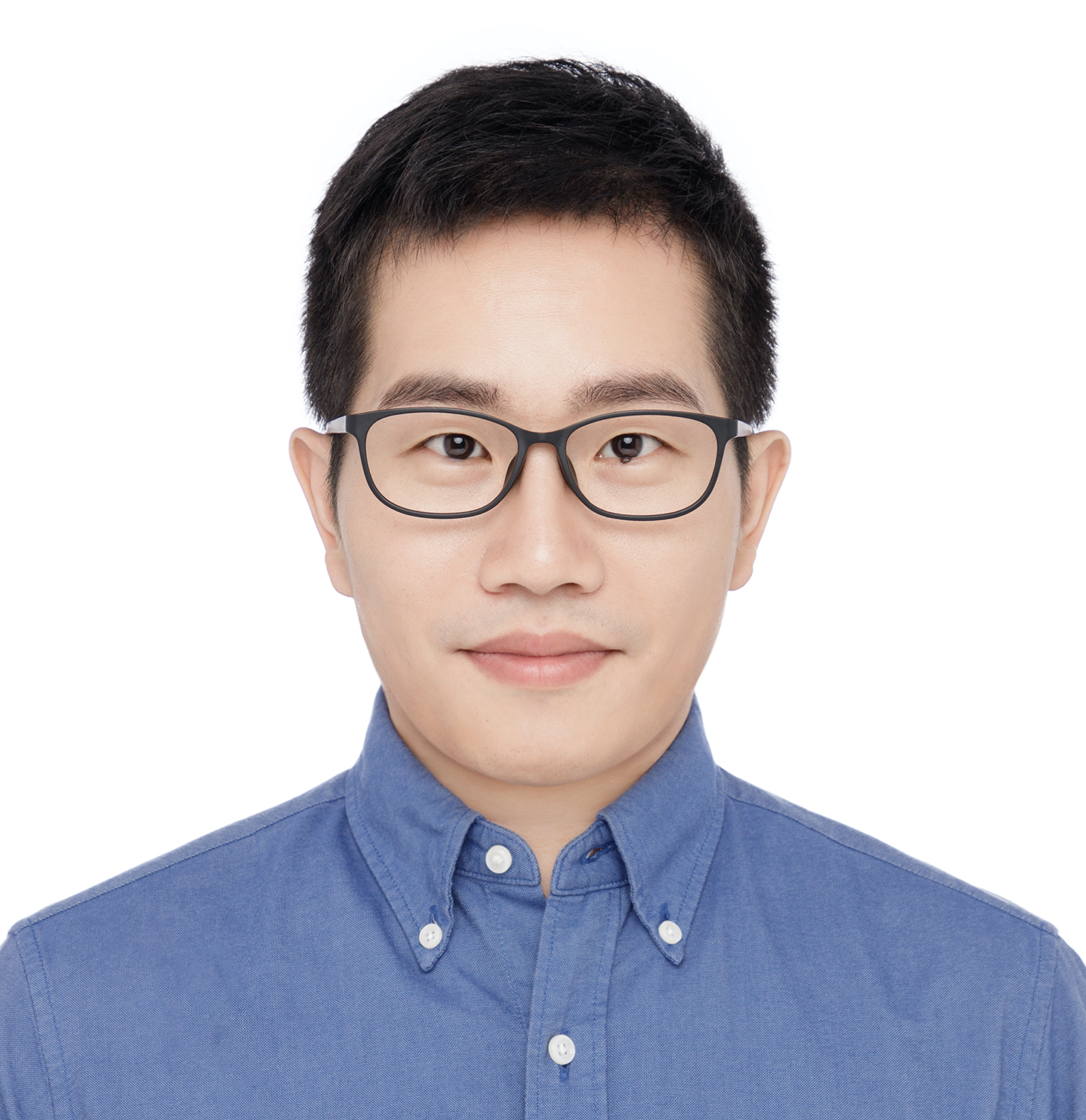}}]{Guanbin Li}
{Guanbin Li}(M'15) is currently an associate professor in School of Data and Computer Science, Sun Yat-sen University. He received his PhD degree from the University of Hong Kong in 2016. His current research interests include computer vision, image processing, and deep learning. He is a recipient of ICCV 2019 Best Paper Nomination Award. He has authorized and co-authorized on more than 80 papers in top-tier academic journals and conferences. He serves as an area chair for the conference of VISAPP. He has been serving as a reviewer for numerous academic journals and conferences such as TPAMI, IJCV, TIP, TMM, TCyb, CVPR, ICCV, ECCV and NeurIPS.
\end{IEEEbiography}

\begin{IEEEbiography}[{\includegraphics[width=1in,height=1.25in,clip,keepaspectratio]{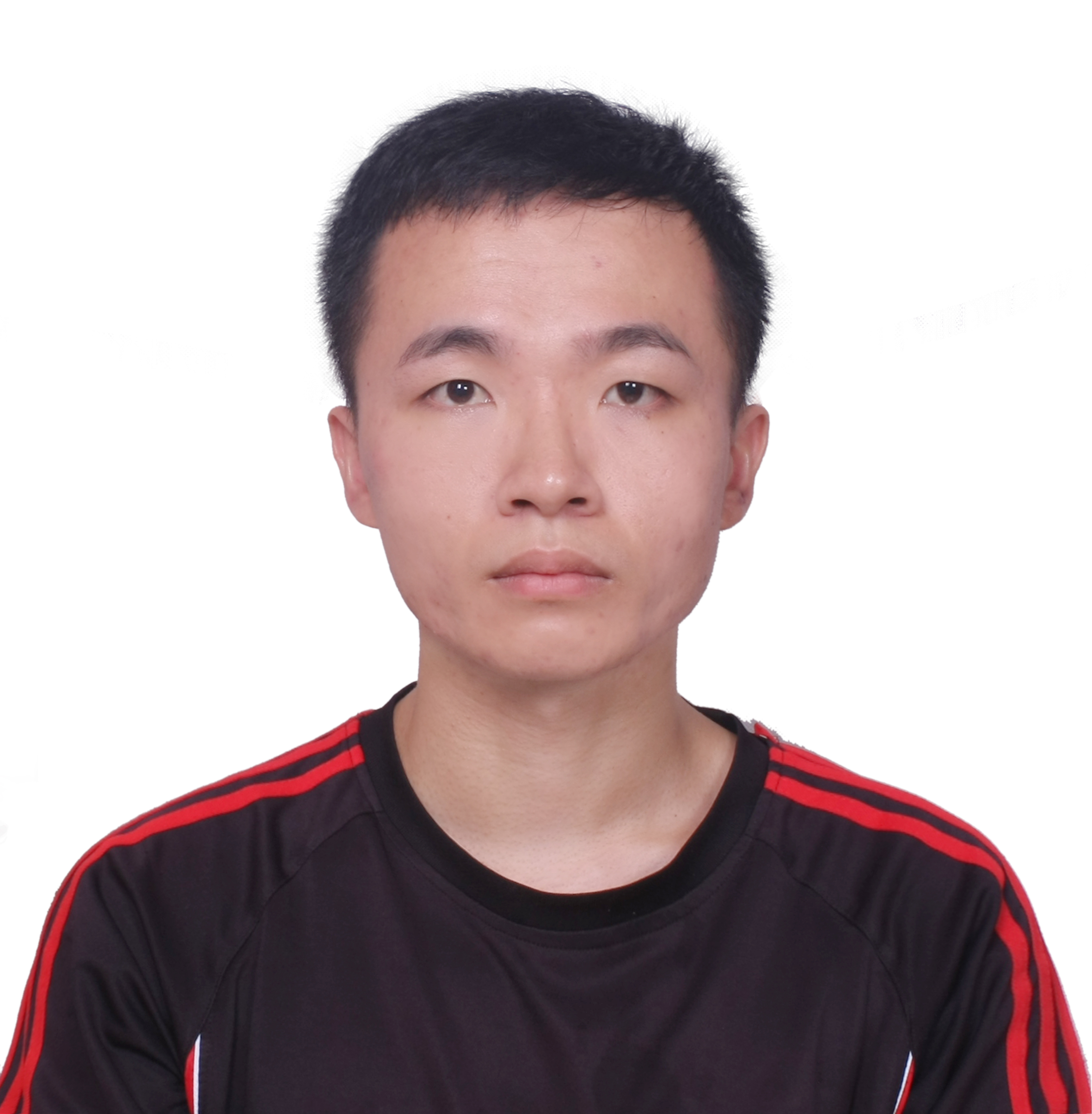}}]{Ziyi Wu}
received the B.E. degree from the School of Computer Science and Engineering, Sun Yat-sen University, Guangzhou, China, in 2020, where he is currently pursuing the Master's degree in computer science. His current research interests include salient object detection and self-supervised learning.
\end{IEEEbiography}

\begin{IEEEbiography}[{\includegraphics[width=1in,height=1.25in,clip,keepaspectratio]{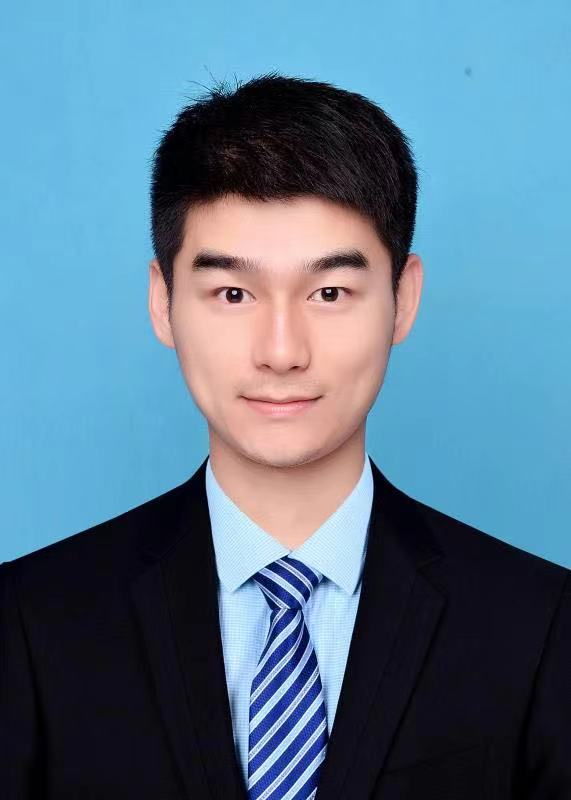}}]{Lei Bai} is a postdoctoral reserach fellow at the School of Electrical and Information Engineering, the University of Sydney, Australia. His reserach interests lie in Machine Learning, Spatial-temporal Learning, and their applicaitons (e.g., Intelligent Transportation, IoT Analytics, and Healthcare). Lei has published a set of peer reviewed papers on top conference and journals such as NeurIPS, CVPR, IJCAI, KDD, Ubicomp, and TITS. He is serving or has served as a program committee member or reviewer for TPAMI, NeurIPS, ICLR, CVPR, ICCV, AAAI, IJCAI, ACM Transactions on Sensor Networks, and so on.
\end{IEEEbiography}

\begin{IEEEbiography}[{\includegraphics[width=1in,height=1.25in,clip,keepaspectratio]{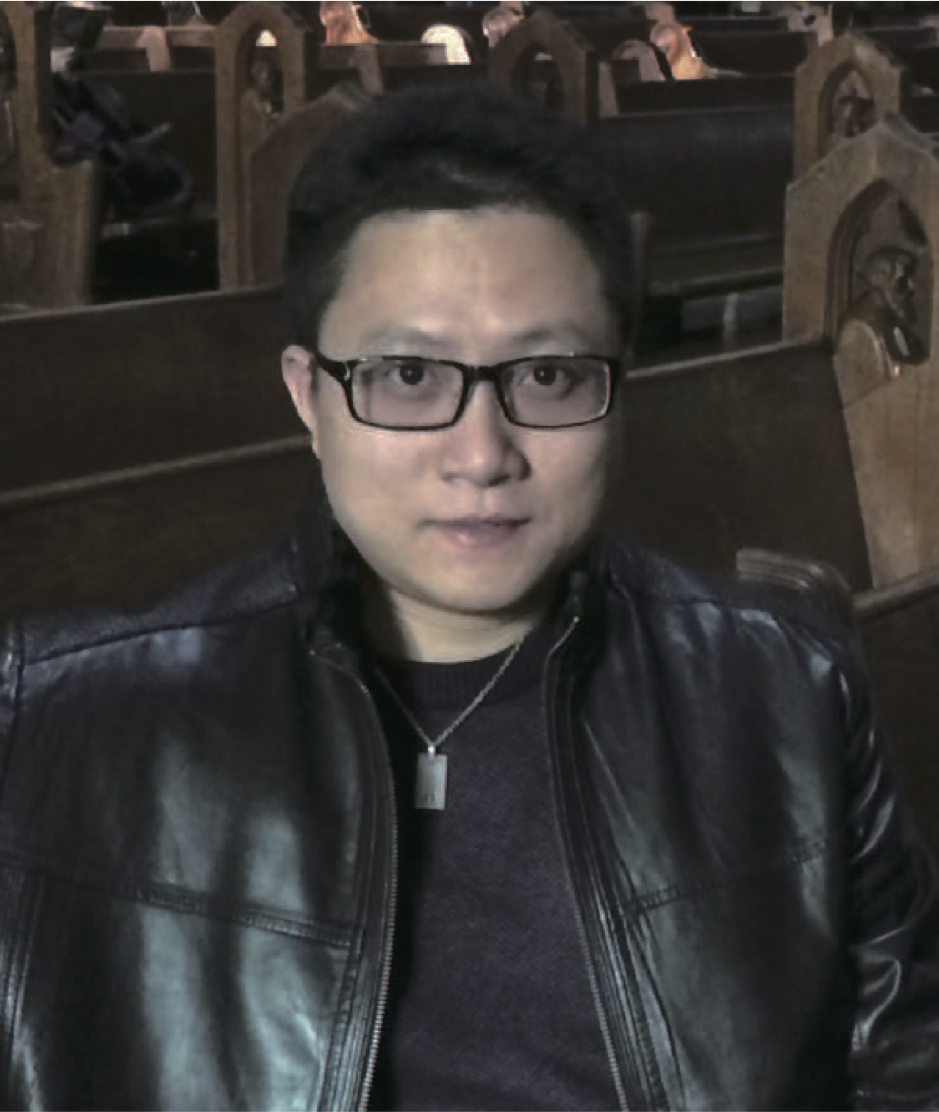}}]{Liang Lin} (M’09, SM’15) is a Full Professor of computer science at Sun Yat-sen University. He served as the Executive Director and Distinguished Scientist of SenseTime Group from 2016 to 2018, leading the R\&D teams for cutting-edge technology transferring. He has authored or co-authored more than 200 papers in leading academic journals and conferences, and his papers have been cited by more than 16,000 times. He is an associate editor of IEEE Trans. Neural Networks and Learning Systems and IEEE Trans. Human-Machine Systems, and served as Area Chairs for numerous conferences such as CVPR, ICCV, SIGKDD and AAAI. He is the recipient of numerous awards and honors including Wu Wen-Jun Artificial Intelligence Award, the First Prize of China Society of Image and Graphics, ICCV Best Paper Nomination in 2019, Annual Best Paper Award by Pattern Recognition (Elsevier) in 2018, Best Paper Dimond Award in IEEE ICME 2017, Google Faculty Award in 2012. His supervised PhD students received ACM China Doctoral Dissertation Award, CCF Best Doctoral Dissertation and CAAI Best Doctoral Dissertation. He is a Fellow of IET.
\end{IEEEbiography}

\end{document}